\documentclass[final,5p,times]{elsarticle}

\usepackage{amssymb}

\usepackage{multirow}
\usepackage{amssymb}
\usepackage[switch]{lineno}

\usepackage[dvipsnames]{xcolor}
\usepackage{subfig}

\usepackage{times}
\usepackage{epsfig}
\usepackage{graphicx}
\usepackage{amsmath}
\usepackage{amssymb}
\usepackage{bbm}
\usepackage{subfig}
\usepackage{pifont}
\usepackage{multirow}

\newcommand{\cmark}{\ding{51}}%
\newcommand{\xmark}{\ding{53}}%

\newcommand{\pairoo}{(\phi_{old},\phi_{old})}
\newcommand{\pairno}{(\phi_{new},\phi_{old})}

\newcommand{\pairnn}{(\phi_{new},\phi_{new})}

\journal{Pattern Recognition}

\begin{document}
\begin{frontmatter}



\title{Dynamic Gradient Reactivation for Backward Compatible Person Re-identification}
\author[label1]{Xiao Pan}
\ead{xiaopan@zju.edu.cn}

\author[label1]{ Hao Luo}
\ead{haoluocsc@zju.edu.cn}

\author[label3]{Weihua Chen}
\ead{kugang.cwh@alibaba-inc.com}

\author[label3]{Fan Wang}
\ead{fan.w@alibaba-inc.com}

\author[label3]{Hao Li}
\ead{lihao.lh@alibaba-inc.com}

\author[label1]{ Wei Jiang\corref{cor1}}
\cortext[cor1]{Corresponding author.}
\ead{jiangweizju@zju.edu.cn}

\author[label1]{ Jianming Zhang}
\ead{ncsl@zju.edu.cn}

\author[label1]{ Jianyang Gu}
\ead{gujianyang@zju.edu.cn}

\author[label2]{Peike Li}
\ead{peike.li@student.uts.edu.au}

\address[label1]{Zhejiang University, Hangzhou, China}
\address[label3]{Alibaba Group, Hangzhou, China}
\address[label2]{University of Technology Sydney, Sydney, Australia}







\begin{abstract}
We study the backward compatible problem for person re-identification (Re-ID), which aims to constrain the  features of an updated new model to be comparable with the existing features from the old model in galleries. Most of the existing works adopt distillation-based methods, which focus on pushing new features to imitate the distribution of the old ones. However, the distillation-based  methods are intrinsically sub-optimal since it forces the new feature space to imitate the inferior old feature space.
 To address this issue, we propose the Ranking-based Backward Compatible Learning (RBCL), which directly optimizes the ranking metric between new features and old features. Different from previous methods, RBCL only pushes the new features to find best-ranking positions in the old feature space instead of strictly alignment, and is in line with the ultimate goal of backward retrieval. 
 However, the sharp sigmoid function used to make the ranking metric differentiable also incurs the gradient vanish issue, therefore stems the ranking refinement during the later period of training. To address this issue, we propose the Dynamic Gradient Reactivation (DGR), which can reactivate the suppressed gradients by adding dynamic computed constant during forward step. To further help targeting the best-ranking positions, we include the Neighbor Context Agents (NCAs) to approximate the entire old feature space during training.
  Unlike previous works which only test on the in-domain settings, we make the first attempt to introduce the cross-domain settings (including both supervised and unsupervised), which are more meaningful and difficult.
 The experimental results on all five settings show that the proposed RBCL outperforms previous state-of-the-art methods by large margins under all settings.  
\end{abstract}




\end{frontmatter}


\section{Introduction}

 Given an query image of a person, person re-identification (Re-ID) retrieves images of the same person from a large gallery of images collected across multiple cameras. In real-world applications, such a gallery is often accumulated over time through pedestrian detection in real-time video streams, with person images represented by learned high-dimensional features. 
 In practice, as the performance of old model becomes worse due to the enlarged gallery, model update is needed. However, it is intractable to recompute the features for all existing images given the sheer size of galleries (billions of images at least), while without recomputing the features, features from different versions of model are in different feature spaces, leading to poor retrieval performance.

\begin{figure}[!t]
\centering
\includegraphics[width=3in]{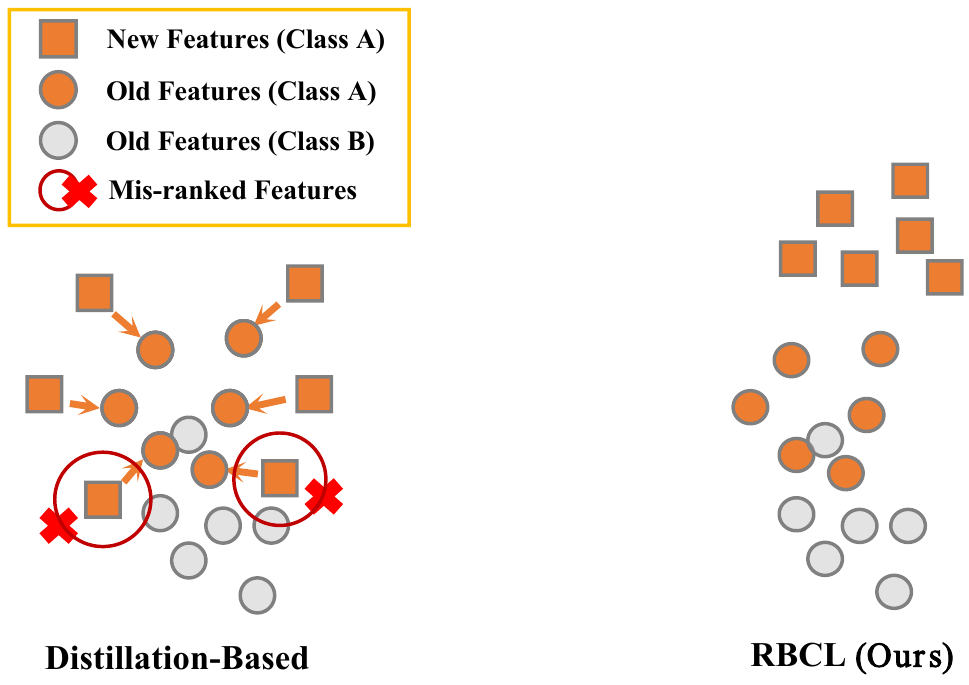}
\caption{Illustration of the difference between the previous distillation-based methods and RBCL (ours). Squares and circles represent the new features and the old features, respectively. Different colors represent different IDs and the arrows represent the optimization direction. 
The distillation-based methods lead to some mis-ranked new features after BCT due to the \textbf{entangled} old feature clusters. 
With the help of RBCL, the new features are pushed into proper ranking positions.}
\label{fig_ranking_and_distillation}
\vspace{-1em}
\end{figure}

 To tackle this problem, the ideal solution is to perform \textbf{Backward Compatible Training (BCT)} \cite{BCT_shen2020towards} for the new model, which targets at making the new features comparable with the existing old features via adding specially designed losses during the training of new model. There has been some works along this direction \cite{BCT_shen2020towards, bct_wang2020unified, bct_chen2019r3}. For example, the Influence Loss \cite{BCT_shen2020towards} conducts the distillation between the new features and the old ones with the help of the old classifier.  RBT \cite{bct_wang2020unified} first maps new features and old ones into a common feature space and then conduct distillation. R3AN \cite{bct_chen2019r3} transforms the new features into the old feature spaces with the reconstruction of the face image. 
 

To our best knowledge, the existing methods are mainly distillation-based. Such methods aim to imitate the distribution of the old model that is less discriminative. However, we argue that they are intrinsically sub-optimal for BCT Re-ID due to the following two disadvantages: (1)  Forcing the new model to imitate the old model would be harmful to the performance of new model itself, especially for the cross-domain setting where the performance of old model on new domain is extremely poor. (2) Due to the low performance of old model, there may exist \emph{entangled clusters}  in the old feature space. The roughly alignment will lead to mis-ranked new features, which will damage the ultimate cross-model retrieval performance.


To address these issues, we propose the \textbf{Ranking-based Backward Compatible Lraining (RBCL)}, which directly optimizes the ranking metric between new features and old features. As illustrated in  Fig. \ref{fig_ranking_and_distillation}, RBCL aims to push new features into best-ranking positions in the old feature space, instead of aligned strictly with it. In this case, the new feature space will not be damaged too much, and the optimization is consistent with the optimal goal of BCT Re-ID, which is to perform retrieval between new features and old features. 
However, to make the discrete ranking metric differentiable, 
the sharp sigmoid function is needed to smooth the indication function, yet also incurs the gradient vanish issue of triplets due to its narrow gradient-effective interval, and therefore stems the refinement of ranking positions during the later period of training.  
To relieve this issue, we propose the \textbf{Dynamic Gradient Reactivation (DGR)}, which can reactivate the suppressed gradients via adding dynamic computed compensate constant during the forward step. With DGR, the gradients for ranking optimization are exponentially enlarged, so that the new features are refined to better-ranking positions. To further target the best-ranking positions, we propose to include the \emph{Neighbor Context Agents (NCAs)} during optimization. NCAs can approximate the entire old feature space with several representative samples, and can provide more triplets for DGR to reactivate.

We evaluate our method under five challenging settings. In contrast to previous works which only 
test on in-domain settings, \textbf{ we make the first attempt to study BCT under cross-domain settings}, including supervised and unsupervised for Re-ID. We emphasize that cross-domain Re-ID  \cite{UDA_fu2019self, UDA_ge2020MMT, UDA_ge2020SPL, UDA_song2020unsupervised, UDA_ADCLUSTER_zhai2020ad} is important for real-world applications but has not been taken into discussion in previous BCT works due to its challenging nature. The experimental results show that our method can achieve superior performance than the existing methods under both in-domain and cross-domain settings.

The main contributions of this paper can be summarized into the following aspects:

\begin{itemize}
\item To the best of our knowledge, we are the first work to study the backward compatible training for person Re-ID under both in-domain and cross-domain settings (including both supervised and unsupervised).
\item Unlike previous distillation-based methods, we propose the RBCL to fully optimize the ranking metric between new features and old features, which leverages DGR to reactivate the suppressed gradients and includes NCAs to further target the best-ranking positions.


\item We perform extensive experiments to demonstrate the effectiveness and superiority of our proposed RBCL.

\end{itemize}

    \begin{figure*}[!t]
    \centering
    
    \includegraphics[width=7.3in]{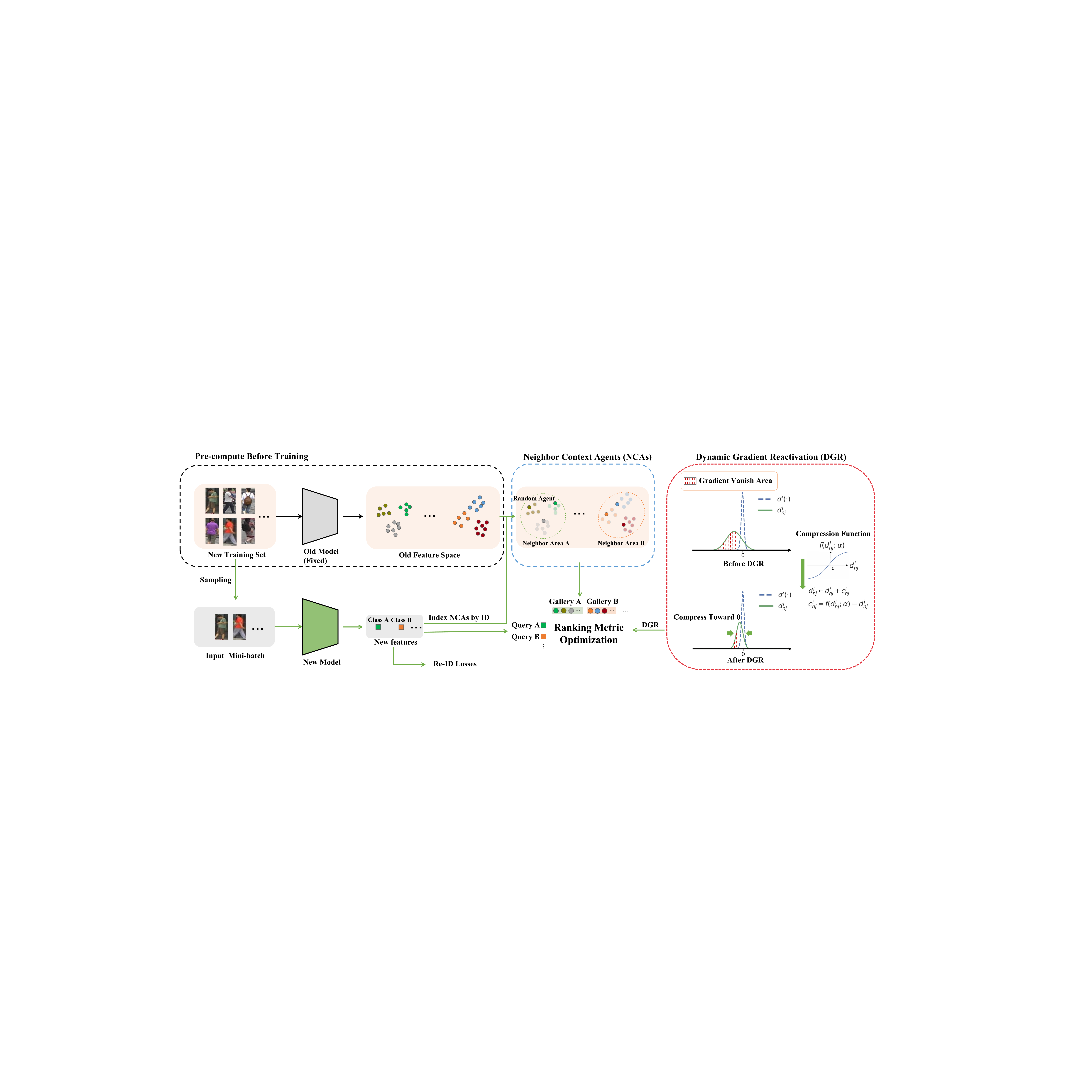}%
    
     
    \caption{ The training pipeline of RBCL. The fixed old features space is pre-computed before iteration. Then, during each iteration, the mini-batch is sent to the new model to get the corresponding new features. After that, the new features are taken as query features and their corresponding NCAs are taken as gallery features for ranking metric optimization. DGR is added during ranking metric optimization to refine the ranking positions. During inference stage, the query features extracted by the new model are retrieved directly with the gallery features from the old model.   }

    \label{fig_pipeline}
    \end{figure*}
\section{Related Work}

    \subsection{Optimizing Ranking for Re-ID}
     
    Multiple works in this community~\cite{ye2016personranking, hermans2017defense,DML_wang2019rankedlistloss,DML_memorybankwang2020cross} have proved that ranking information is important for Re-ID as a retrieval task.
    The typical methods for optimizing the ranking for Re-ID are Deep Metric Learning (DML) methods, which intend to optimize the distance metric.
    Generally, the DML methods can be further divided into pairwise \cite{DML_Contrastive_oh2016deep,DML_Contrastive_schroff2015facenet} and triplet-wise methods \cite{hermans2017defense, DML_N_pair_sohn2016improved, DML_lift_structure_oh2016deep, DML_proxy_NCA_movshovitz2017no, DML_wang2019rankedlistloss, DML_memorybankwang2020cross}.
     To better exploit the information when comparing features in a mini-batch, hard example mining is first combined with Triplet loss by \cite{hermans2017defense} and achieves significant improvement.
    To get richer pairwise information beyond a mini-batch, Wang \textit{et al.} \cite{DML_memorybankwang2020cross}  propose the cross-batch memory (XBM) mechanism which memorizes the features of the past iterations. 
     

    Another line of works for the ranking optimization is to directly optimize the Average Precision (AP), which belongs to the ranking metric optimization. Average Precision (AP) is an important metric for person Re-ID. The main problem for directly optimizing AP is that the calculation of AP includes a discrete ranking function, which is neither differentiable nor decomposable. Brown \textit{et al.} \cite{DML_brown2020smooth} propose the Smooth-AP which optimizes a smoothed approximation of AP and show its satisfactory performance on large-scale datasets.

    The improvement from these ranking-motivated loss functions shows that exploiting the ranking is crucial for the Re-ID task.  Our proposed RBCL also takes the ranking into consideration, which has been ignored in previous works for BCT. We follow the ranking metric optimization \cite{DML_brown2020smooth} since the distance metric optimization is sub-optimal when evaluating using a ranking metric. 
     
    
\subsection{Backward Compatible Training}

     Chen \textit{et al.} \cite{bct_chen2019r3} first formalize the cross-model face recognition task and propose the R3AN to transfer the features of the query model to the feature space of the gallery model. R3AN is specialized for the face recognition task and requires the two models to be similar, which is not applicable in some real-world model upgrade scenario. Wang \textit{et al.} \cite{bct_wang2020unified} propose to map the features of the two models into a unified feature space through the proposed Residual Bottleneck Transformation (RBT) blocks. Then, several distillation (imitation)-style losses are applied in the unified feature space. Both R3AN and RBT require extra transformations on features, which brings about extra computation and is intractable in real-world applications. Shen \textit{et al.} \cite{BCT_shen2020towards} bring up the term of Backward Compatible Training (BCT) for the first time, which focus more on the model upgrade scenario compared with the above mentioned two methods. They propose the Influence Loss, which performs distillation between the logits of new features and old ones through the classifier of the old model, and the classification loss of the new model is combined with the Influence Loss when training the new model. 
     This form of combination is similar to Learning Without Forgetting (LWF) \cite{incremental_li2017Lwf} which is the typical solution of incremental learning, but differs in that the classifier of the old model is fixed here. 
     Meng \textit{et al.} \cite{meng2021lce} propose LCE, which aligns the centers between new features and old features, and then restrict more compact intra-class distributions for new features. Although LCE won't distill the new feature distribution to the old one strictly in instance-level, it still aligns these two distributions, which is ineffective for handling the entangled old clusters. Different from the previous methods, our proposed RBCL directly optimizes the ranking metric between new features and old features.
         
     


\section{Methodology}
\label{sec:method}
In this section, we first introduce the backward compatible training in formula and then introduce our proposed Ranking-based Backward Compatible Learning (RBCL). 

\subsection{Problem Formalization}
\label{sec:problem_formalization}
Assuming that a new model trained without any consideration of BCT is denoted as $\phi_{new}$, and the fixed old model is denoted as $\phi_{old}$. To improve the backward compatibility, we design a method  $L$ and integrate it into the training process of $\phi_{new}$. We then name the new model trained with the help of method $L$ as $\phi_{new-L}$. 

\begin{sloppypar}
For a test set $\mathcal{D}_{test}$, its evaluation metric (e.g., mAP or Rank-1) is represented as $M(\phi_{q},\phi_{g};\mathcal{D}_{test})$, where the query features are extracted by $\phi_{q}$, and gallery features are extracted by $\phi_{g}$. Since $\phi_{new}$ and $\phi_{old}$ are in different feature spaces, directly evaluating between $\phi_{new}$ and $\phi_{old}$, i.e., $M(\phi_{new}, \phi_{old}; \mathcal{D}_{test})$, gives unsatisfactory performance. Therefore, our goal is to design a method $L$, which can make $ M(\phi_{new-L}, \phi_{old}; \mathcal{D}_{test})  >  M(\phi_{old},\phi_{old}; \mathcal{D}_{test})$ as much as possible.
\end{sloppypar}

\subsection{Ranking-based Backward Compatible Learning}
The pipeline of our RBCL is illustrated in Fig. \ref{fig_pipeline}. Before iteration, the training set of the new model is first sent to the old model to get the fixed old feature space. Then, during each iteration, a mini-batch is sampled from the training set, and then sent to the new model to get the corresponding new features. After that, the new features are taken as the query features and their corresponding NCAs (indexed by IDs from the pre-computed old feature space) are taken as the gallery features. Finally, the metric optimization is conducted between query features and gallery features with the help of DGR, and the new features are sent to Re-ID losses for discriminative learning. During inference, the the query features are extracted by the new model, and then directly retrieved (cosine similarity)  with the old gallery features. 
    
    \subsubsection{Ranking Metric Optimization}
    Inspired by \cite{DML_brown2020smooth}, we use the smoothed mean Average Precision (mAP) metric to optimize the ranking between new features and old features during training. Given a query feature $f_{i}$ from query feature set $F_q$, which is extracted by the new model, the gallery feature set $F_{g}$ extracted by the old model is split into positive set $\mathbbm{P_{i}}$ and negative set $\mathbbm{N_{i}}$, which are formed by all instances of the same class and of different classes, respectively. 
    Then, the smoothed AP for query $i$ becomes:
    \begin{equation}
    \label{eq:smoothAP}
    \small
        AP_{i} \! = \!  \frac{1}{|\mathbbm{P}_i|} \! \sum_{j \in \mathbbm{P}_{i}}  \!
    \frac{1 \! +  \! \sum_{p \in \mathbbm{P}_i \backslash \{j\} } \sigma(d_{pj}^i)   }
    { 1 + \sum_{p \in \mathbbm{P}_i \backslash \{j\}} \sigma( d_{pj}^i) \!  +  \! \sum_{n \in \mathbbm{N}_i}  \sigma(d_{nj}^i) }, 
    \end{equation}
    where $d_{pj}^i = s_{ip} - s_{ij}$ and $d_{nj}^i = s_{in} - s_{ij}$, in which $s_{ij}$ represents for the cosine similarity between $f_i$ and $f_j$; and $\sigma(\cdot)$ represents for the sigmoid function:
    \begin{equation}
        \label{simoid}
         \sigma = \frac{1}{1+e^{-x/\tau}}.
    \end{equation}
    which is the relaxed form of indication function to make AP differentiable \cite{DML_brown2020smooth}. Then, the final smoothed mean Average Precision (mAP) loss is:
    \begin{equation}
        \mathcal{L}_{m} =1-\frac{1}{|F_{q}|}\sum_{i \in |F_q|}AP_{i}.
        \label{smoothap}
    \end{equation}
    
    \subsubsection{Dynamic Gradient Reactivation}
    
    \begin{figure}[!t]
\centering
\subfloat[]{\includegraphics[width=1.72in]{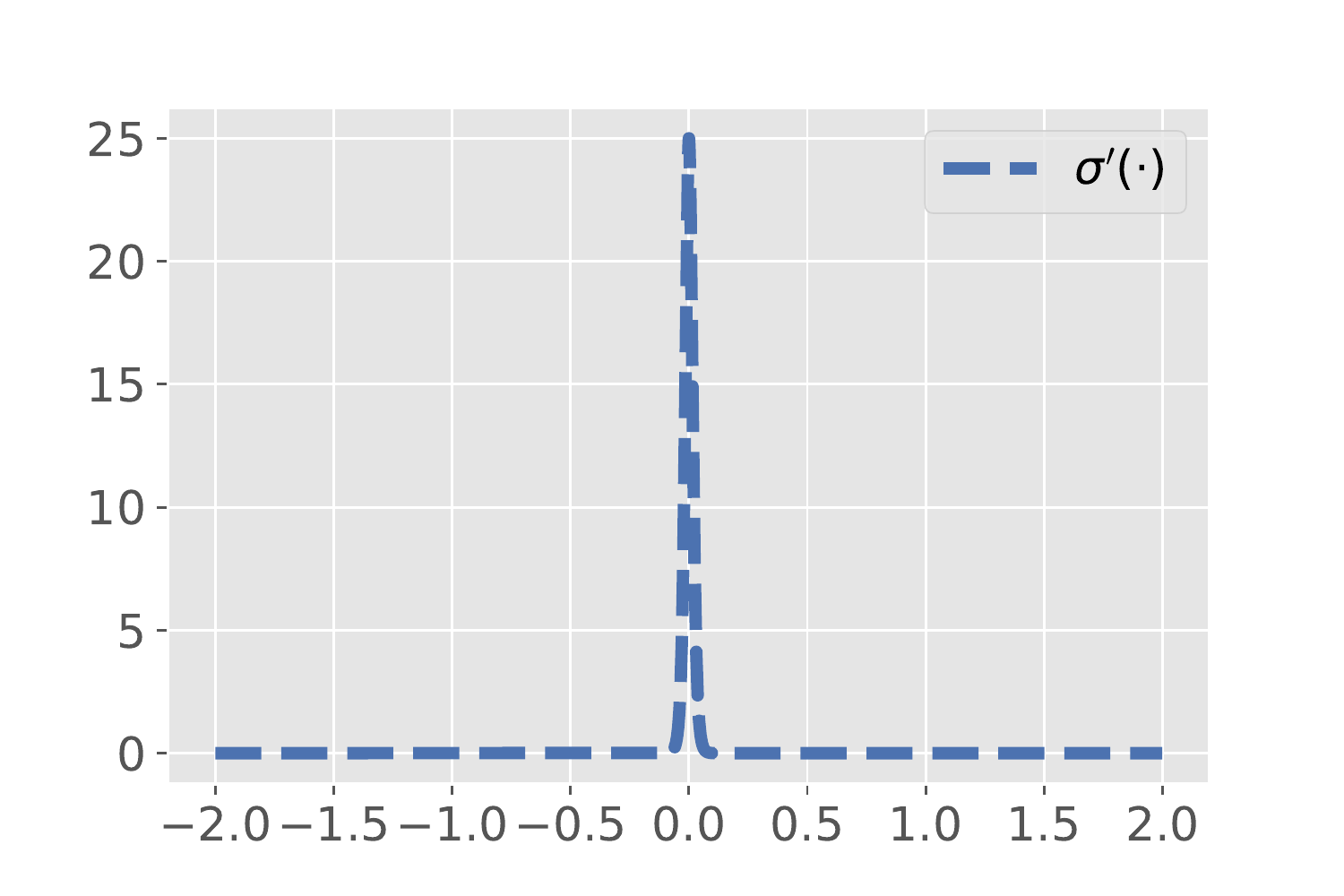}%
\label{fig_derivative_of_simoid}}
\subfloat[]{\includegraphics[width=1.80in]{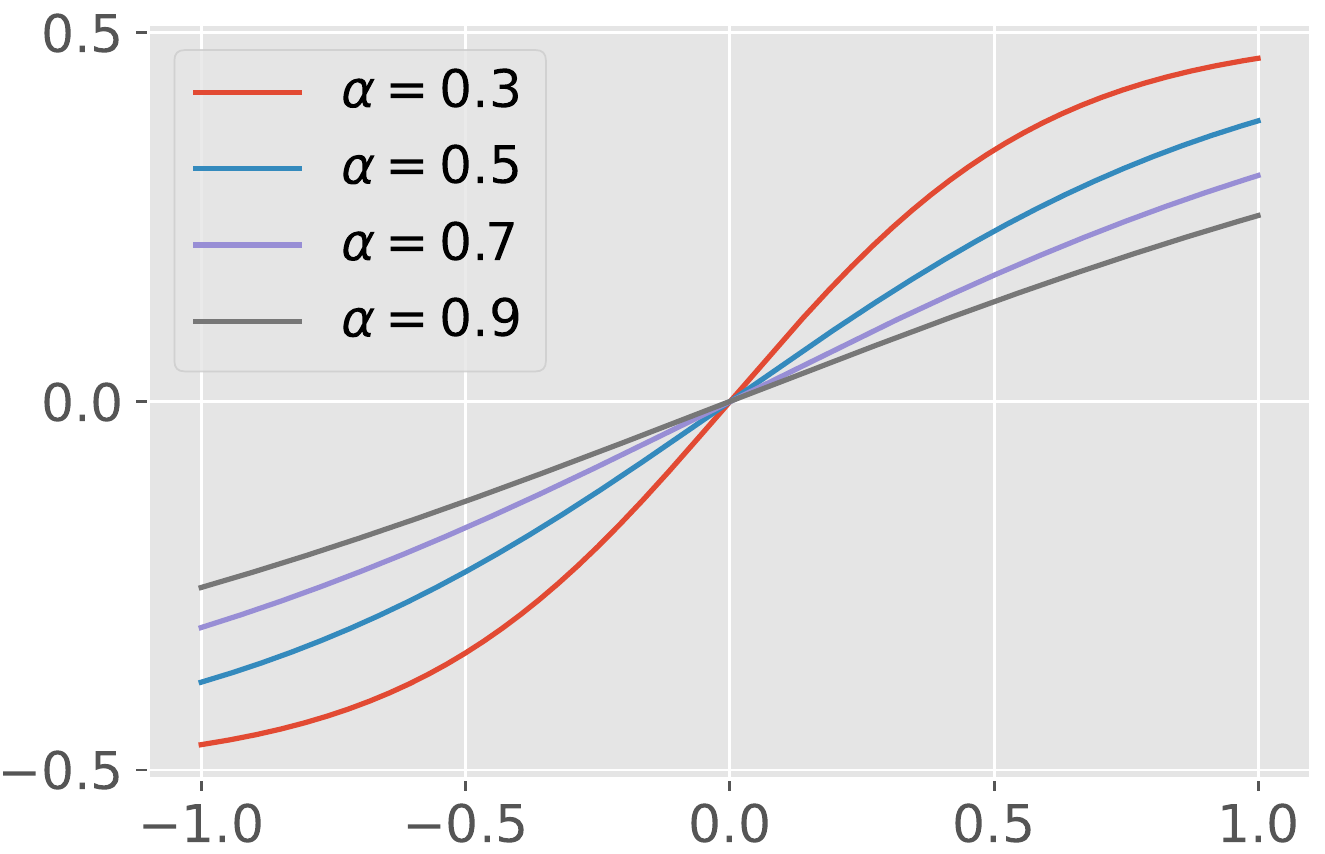}%
\label{fig_influence_of_alpha}}
\caption{ (a) The derivative of $\sigma(\cdot)$ when $\tau = 0.01$. (b) The influence of $\alpha$ to the compression function (Eq. \ref{eq:compression fuction}).}

\end{figure}

    The main goal of optimizing $\mathcal{L}_{m}$ is to make $AP_i$ close to 1, thus, although not completely equivalent, the key is to minimize the $\sum_{n \in \mathbbm{N}_i}  \sigma(d_{nj}^i)$ term in the denominator of Eq. \ref{eq:smoothAP}. $d_{nj}^i = s_{in} - s_{ij}$ is a triplet composed of an anchor $i$, a positive sample $j$, and a negative sample $n$. Minimizing $\sum_{n \in \mathbbm{N}_i}  \sigma(d_{nj}^i)$ means to decrease the similarity of negative pairs and increase the similarity of positive pairs, i.e., to optimize the ranking. Therefore, the gradient of each triplet $d_{nj}^i$ plays an important role in the ranking optimization. However, $d_{nj}^i$ is wrapped by the sigmoid function $\sigma'(\cdot)$, whose gradient-effective interval is quite narrow since the derivative decreases exponentially as the distance from $0$ becomes larger (see Fig. \ref{fig_derivative_of_simoid}), especially when $\tau$ is small ($\tau=0.01$ performs best as mentioned in \cite{DML_brown2020smooth}). In this case, only the triplets close to $0$ can achieve effective gradients, while for those away from $0$ with certain distance, the gradients are extremely small, i.e., vanished. 
    
    On the other hand, during the later period of training, we have two observations about the distribution of $d_{nj}^i$ (see the green line in the upper right corner of Fig. \ref{fig_pipeline}): (1) Most of the triplets are in proper ranking order after considerable time of optimization, i.e., the majority of $d_{nj}^i$ are smaller than $0$ with certain distance, and therefore achieve extremely small gradients. However, we propose that there still exists valuable information among them which can further help refining the ranking positions. Intuitively, the best-ranking positions in the fixed old feature space are more likely to be achieved when more triplets (ranking relations) are considered. (2) There still exist several extremely hard negative samples which are closer to the anchors than the positive samples with certain distance due to the entangled clusters in the old feature space, i.e., several $d_{nj}^i$ are still larger than $0$ with certain distance. However, such extremely hard negative samples can hardly be optimized due to the vanished gradients. Therefore, their gradients should also be enhanced.  Based on (1) and (2), we conclude that the gradient vanish issue stems the further ranking optimization during the later period of training. 
    
    To address this issue, we propose to reactivate these triplets by compressing their distribution toward the gradient-effective interval of $\sigma'(\cdot)$ via adding a reactivation constant $c_{nj}^i$ during the forward step:
    \begin{equation}
    \begin{aligned} 
        d_{nj}^i &\xleftarrow[]{}{} d_{nj}^i + c_{nj}^i, \\
        c_{nj}^i &= f(d_{nj}^i;\alpha) - d_{nj}^i,
    \end{aligned}
    \end{equation}
    where $f(d_{nj}^i;\alpha)$ is a compression function, and $\alpha$ controls the degree of compression. Specifically, we choose $f(d_{nj}^i;\alpha)$ as:
    \begin{equation}
    \label{eq:compression fuction}
        f(d_{nj}^i;\alpha) = \sigma(d_{nj}^i;\alpha) - 0.5,
    \end{equation}
    where $\alpha$ is the anneal of sigmoid function. The influence of $\alpha$ is illustrated in Fig. \ref{fig_influence_of_alpha}. The larger the $\alpha$ is, the greater compression is. Note that $c_{nj}^i$ is a constant without gradient, and is customized for each $d_{nj}^i$, therefore we term it as ``dynamic''. 
    
    As illustrated in the bottom right corner of Fig. \ref{fig_pipeline}, the distribution of the triplets (${d_{nj}^i}$) is compressed toward $0$ after adding DGR, therefore their gradients are exponentially enhanced in general, which promotes refining the ranking positions. 
    
    \subsubsection{Neighbor Context Agents}
     Although DGR can refine the ranking positions by enhancing gradients, the ranking optimization still lacks the global perspective for targeting the best-ranking positions. Intuitively, the global best-ranking positions can be achieved only when the model sees the entire old feature space. However, the computation overhead will be extremely expensive, especially when the training set is large. Thus, we propose to include Neighbor Context Agents (NCAs) during training to approximate the entire old feature space. As illustrated in the middle of Fig. \ref{fig_pipeline}, NCAs are composed of random agents from each class in the neighbor area of current mini-batch. 
     More concretely, we pre-compute the old features of the training set, and then build an adjacency matrix based on the euclidean distance between cluster centroids. 
     Then,  during each iteration, given the samples from the $c$-th class in the mini-batch, we choose the $K$ classes closest to the $c$-th class in the old feature space as its neighbor set $\mathcal{N}_c$. For each class in $\mathcal{N}_c \cup \{c\}$, we randomly pick one sample as the agent at each iteration: 
     \begin{equation}
         A_{old}^c = \{ Rand (F_{old}^k) | k  \in \mathcal{N}_c \cup \{c\}    \} ,
     \end{equation}
     where $Rand(\cdot)$ represents for the operation of randomly selecting one feature, $F_{old}^k$ represents the set of old features for $k$-th class, and $A_{old}^c$ represents the set of the selected agents for $c$-th class. Assuming that there are totally $B$ classes in a mini-batch (\{$k_1$, $k_2$, ..., $k_B \}$), all the gallery features picked at each iteration are:
     \begin{equation}
         F_{g} = A_{old}^{k_1} \cup A_{old}^{k_2} \cup ... \cup A_{old}^{k_B} .
     \end{equation}
         
     After training with abundant iterations, the NCAs will walk through the entire old feature space, which can help targeting the best-ranking positions. Also, NCAs can provide more triplets for DGR to reactivate.

    \subsubsection{Overall Losses}
    We train  
    $\mathcal{L}_{m}$ together with the Re-ID losses. Formally, the final loss function for backward compatible training is:
    \begin{equation}
        \mathcal{L}_{total} = \mathcal{L}_{tri} + \mathcal{L}_{id} + \mathcal{L}_{m} , 
    \end{equation}
    where $\mathcal{L}_{tri}$ and $\mathcal{L}_{id}$ represent for the Hard Mining Triplet Loss \cite{hermans2017defense} and ID Loss \cite{zheng2017discriminatively}, which are commonly used losses in Re-ID for discriminative learning.

\section{Experimental Setting Formalization}
\begin{table}[t]
    \footnotesize
    \setlength\tabcolsep{4.2pt}
    \centering
    
    \begin{tabular}{c|l|ccccc|c|c|cl}
    \cline{1-10}
    \multicolumn{2}{l|}{}         & \multicolumn{1}{l}{$\mathcal{T}_{old}$} & \multicolumn{1}{l}{$\mathcal{B}_{old}$} & $\mathcal{T}_{new}$ & $\mathcal{B}_{new}$ & \multicolumn{1}{l|}{$\mathcal{D}_{test}$} & \multicolumn{1}{l|}{Sup}        & \multicolumn{1}{l|}{CD}       & \multicolumn{1}{l}{NS}      &                      \\ \cline{1-10}
    \multirow{2}{*}{ID-S-1} & M2M & $\mathcal{T}_{M0.1}$                    & $R_{50}$                      & $\mathcal{T}_{M}$   & $R_{50}$  & $\mathcal{D}_{M}$                         & \multirow{2}{*}{\cmark} & \multirow{2}{*}{\xmark} & \multirow{2}{*}{\xmark} &                      \\
                            & D2D & $\mathcal{T}_{D0.1}$                    & $R_{50}$                      & $\mathcal{T}_{D}$   & $R_{50}$  & $\mathcal{D}_{D}$                         &                                        &                                        &                                        &                      \\ \cline{1-10}
    \multirow{2}{*}{ID-S-2} & M2M & $\mathcal{T}_{M0.1}$                    & $R_{50}$                      & $\mathcal{T}_{M}$   & $R_{101ibn}$ & $\mathcal{D}_{M}$                         & \multirow{2}{*}{\cmark} & \multirow{2}{*}{\xmark} & \multirow{2}{*}{\cmark} &                      \\
                            & D2D & $\mathcal{T}_{D0.1}$                    & $R_{50}$                      & $\mathcal{T}_{D}$   & $R_{101ibn}$ & $\mathcal{D}_{D}$                         &                                        &                                        &                                        &                      \\ \cline{1-10}
    \multirow{2}{*}{CD-S-1} & M2D & $\mathcal{T}_{M}$                       & $R_{50}$                      & $\mathcal{T}_{D}$   & $R_{50}$  & $\mathcal{D}_{D}$                         & \multirow{2}{*}{\cmark} & \multirow{2}{*}{\cmark} & \multirow{2}{*}{\xmark} &                      \\
                            & D2M & $\mathcal{T}_{D}$                       & $R_{50}$                      & $\mathcal{T}_{M}$   & $R_{50}$  & $\mathcal{D}_{M}$                         &                                        &                                        &                                        &                      \\ \cline{1-10}
    \multirow{2}{*}{CD-S-2} & M2D & $\mathcal{T}_{M}$                       & $R_{50}$                      & $\mathcal{T}_{D}$   & $R_{101ibn}$ & $\mathcal{D}_{D}$                         & \multirow{2}{*}{\cmark} & \multirow{2}{*}{\cmark} & \multirow{2}{*}{\cmark} &                      \\
                            & D2M & $\mathcal{T}_{D}$                       & $R_{50}$                      & $\mathcal{T}_{M}$   & $R_{101ibn}$ & $\mathcal{D}_{M}$                         &                                        &                                        &                                        &                      \\ \cline{1-10}
    \multirow{2}{*}{CD-US}  & M2D & $\mathcal{T}_{M}$                       & $R_{50}$                      & $\mathcal{T}_{D}^{u}$   & $R_{50}$  & $\mathcal{D}_{D}$                         & \multirow{2}{*}{\xmark} & \multirow{2}{*}{\cmark} & \multirow{2}{*}{\xmark} & \multicolumn{1}{c}{} \\
                            & D2M & $\mathcal{T}_{D}$                       & $R_{50}$                      & $\mathcal{T}_{M}^{u}$   & $R_{50}$  & $\mathcal{D}_{M}$                         &                                        &                                        &                                        & \multicolumn{1}{c}{} \\ \cline{1-10}
    \end{tabular}

    \caption{ The detailed configurations of each setting. ``Sup'', ``CD'', and ``NS'' represent for ``Supervised'', ``Cross Domain'', and ``New Structure'', separately.  $\mathcal{T}_{M0.1}$  and $\mathcal{T}_{D0.1}$  represent for 10\% of the training set (randomly split by identities). $\mathcal{T}_{M}^{u}$ and $\mathcal{T}_{D}^{u}$ represent for the unlabelled training set.}
    \label{table:detailed_configurations}
    \end{table}
    
    As the backward compatible training for person Re-ID is a relatively new topic,  we would like to first formalize the experimental setup in this section before describing the details of experimental results. 
    
    \subsection{Datasets and Backbones}
  
    \begin{sloppypar} We conduct experiments on Market-1501 \cite{marketzheng2015scalable} and DukeMTMC-reID \cite{dukeristani2016performance,dukezheng2017unlabeled}. Market-1501 contains 12,936 images from 751 identities for training and 19,732 images from 750 identities for testing. The images are captured from 6 cameras. DukeMTMC-reID contains 16,522 images from 702 identities for training and the rest images from 702 identities for testing. ResNet-50 \cite{he2016deepResnet} and IBN-ResNet-101 \cite{ibnpan2018two} are used as the backbones. \end{sloppypar}

    \subsection{Setting Types}
    
    \noindent \textbf{Notations.} Formally, we represent the different backbones as $\mathcal{B}=\{R_{50}, R_{101ibn}\}$, where $R_{50}$ and $R_{101ibn}$ represent for ResNet-50 and IBN-ResNet-101, respectively. The training sets are represented as $\mathcal{T} = \{ \mathcal{T}_{M}, \mathcal{T}_{D} \}$, where $\mathcal{T}_{M}$ represents for the training set of Market1501, and $\mathcal{T}_{D}$ for DukeMTMC-reID. Similarly,  their test sets are denoted as $\mathcal{D}=\{\mathcal{D}_M, \mathcal{D}_D\}$.  The detailed configurations of all the settings below are illustrated in Table. \ref{table:detailed_configurations}.


    \subsubsection{In-Domain Settings}
    
    \begin{sloppypar}
    We first test the BCT performance under the basic in-domain settings. For each in-domain setting, we test on both Market1501 (M2M) and DukeMTMC-reID (D2D).
    \end{sloppypar}
    
    \noindent \textbf{In-Domain Supervised  Setting 1 (ID-S-1)}:  
    Under this setting, we randomly split 10\% IDs from the original training set for the training of old model, and train the new model with the entire training set.

    \noindent \textbf{In-Domain Supervised  Setting 2 (ID-S-2)}: 
    To verify the robustness  of the methods to the structure variance, under this setting, we replace the backbone of the new model in ID-S-1 with IBN-ResNet-101. 
    
    \subsubsection{Cross-Domain Settings}
    Aside from the basic in-domain settings, we also test our methods under challenging cross-domain settings targeting several real-world scenarios. For each cross-domain setting, we test on two directions, i.e., from Market1501 to DukeMTMC-reID (M2D) and from DukeMTMC-reID to Market1501 (D2M). 
    
    \noindent \textbf{Cross-Domain Supervised  Setting 1 (CD-S-1)}: 
    In real-world applications, the poor performance of the old model may be caused by the domain bias between the training set and the application scenario (test set). To improve the performance, new training set from the target domain is collected for model upgrade.
    Moreover, training set for the old model is not accessible any more due to the expensive storage overhead and privacy issue. This setup is challenging due to the domain gap and the poor performance of the old model. 
        
    

    \noindent \textbf{Cross-Domain Supervised  Setting 2 (CD-S-2)}:  Similar to ID-S-2, we use IBN-ResNet-101 as the backbone of new model, while ResNet-50 for the old model under this setting.

    \noindent \textbf{Cross-Domain Unsupervised  Setting (CD-US)}: In real-world applications, achieving the large-scale labelled training set is expensive, thus unsupervised domain adaption (UDA) is extensively studied  \cite{UDA_fu2019self, UDA_ge2020MMT, UDA_ge2020SPL, UDA_song2020unsupervised, UDA_ADCLUSTER_zhai2020ad}. 
    We also prove the effectiveness of our method under this challenging setting. We conduct UDA from the domain of old model to the domain of new model. 

        \begin{table*}[t]
        \centering
        \footnotesize
        \setlength\tabcolsep{4.5pt}
        \begin{tabular}{l|cc|cc|cc|cc|cc|cc|cc|cc|cccc}
        \hline
                                & \multicolumn{4}{c|}{ID-S-1}                         & \multicolumn{4}{c|}{ID-S-2}                         & \multicolumn{4}{c|}{CD-S-1}                         & \multicolumn{4}{c|}{CD-S-2}                         & \multicolumn{4}{c}{CD-US}                                   \\ \hline
        \multirow{2}{*}{Method} & \multicolumn{2}{c|}{M2M} & \multicolumn{2}{c|}{D2D} & \multicolumn{2}{c|}{M2M} & \multicolumn{2}{c|}{D2D} & \multicolumn{2}{c|}{M2D} & \multicolumn{2}{c|}{D2M} & \multicolumn{2}{c|}{M2D} & \multicolumn{2}{c|}{D2M} & \multicolumn{2}{c|}{M2D}          & \multicolumn{2}{c}{D2M} \\ \cline{2-21} 
                                & mAP         & R-1        & mAP         & R-1        & mAP         & R-1        & mAP         & R-1        & mAP         & R-1        & mAP         & R-1        & mAP         & R-1        & mAP         & R-1        & mAP  & \multicolumn{1}{c|}{R-1}  & mAP         & R-1        \\ \hline
        Direct                  & 37.5        & 61.0       & 36.5        & 56.4       & 37.5        & 61.0       & 36.5        & 56.4       & 17.9        & 31.9       & 22.4        & 49.2       & 17.9        & 31.9       & 22.4        & 49.2       & 28.7 & \multicolumn{1}{c|}{44.5} & 30.0        & 58.8       \\
        LB                      & 50.0        & 74.1       & 43.6        & 66.2       & 0.2         & 0.1        & 0.2         & 0.1        & 32.2        & 54.5       & 33.0        & 57.5       & 0.2         & 0.0        & 0.2         & 0.1        & 0.4  & \multicolumn{1}{c|}{0.3}  & 0.2         & 0.1        \\
        UB                      & 85.8        & 93.8       & 76.6        & 86.3       & 89.1        & 95.7       & 81.0        & 90.6       & 76.6        & 86.3       & 85.8        & 93.8       & 81.0        & 90.6       & 89.1        & 95.7       & 58.7 & \multicolumn{1}{c|}{74.5} & 67.6        & 85.0       \\ \hline
        \end{tabular}
        \caption{ The primary experiments of BCT under all the settings. No additional loss for improving the backward compatibility is used. ``Direct'', ``LB'' and ``UB'' represent for $M\pairoo$, $M\pairno$,  and $M\pairnn$ (defined in Sec. \ref{sec:problem_formalization}), separately.}
        \label{table:basic_ablation}
        \end{table*}
        \begin{table*}[t]
        \centering
        \footnotesize
        \setlength\tabcolsep{2.2pt}
        
        \begin{tabular}{l|cc|cc|cc|cc|cc|cc|cc|cc|cccc}
        \hline
                                & \multicolumn{4}{c|}{ID-S-1}                                                                                                                                                                                                                                       & \multicolumn{4}{c|}{ID-S-2}                                                                                                                                                                                                                                       & \multicolumn{4}{c|}{CD-S-1}                                                                                                                                                                                                                                       & \multicolumn{4}{c|}{CD-S-2}                                                                                                                                                                                                                                       & \multicolumn{4}{c}{CD-US}                                                                                                                                                                                                                                                              \\ \hline
        \multirow{2}{*}{Method} & \multicolumn{2}{c|}{M2M}                                                                                                        & \multicolumn{2}{c|}{D2D}                                                                                                        & \multicolumn{2}{c|}{M2M}                                                                                                        & \multicolumn{2}{c|}{D2D}                                                                                                        & \multicolumn{2}{c|}{M2D}                                                                                                        & \multicolumn{2}{c|}{D2M}                                                                                                        & \multicolumn{2}{c|}{M2D}                                                                                                        & \multicolumn{2}{c|}{D2M}                                                                                                        & \multicolumn{2}{c|}{M2D}                                                                                                                              & \multicolumn{2}{c}{D2M}                                                                                                        \\ \cline{2-21} 
                                & mAP                                                            & R-1                                                            & mAP                                                            & R-1                                                            & mAP                                                            & R-1                                                            & mAP                                                            & R-1                                                            & mAP                                                            & R-1                                                            & mAP                                                            & R-1                                                            & mAP                                                            & R-1                                                            & mAP                                                            & R-1                                                            & mAP                                                            & \multicolumn{1}{c|}{R-1}                                                            & mAP                                                            & R-1                                                            \\ \hline
        baseline                & 50.0                                                           & 74.1                                                           & 43.6                                                           & 66.2                                                           & 0.2                                                            & 0.1                                                            & 0.2                                                            & 0.1                                                            & 32.2                                                           & 54.5                                                           & 33.0                                                           & 57.5                                                           & 0.2                                                            & 0.0                                                            & 0.2                                                            & 0.1                                                            & 0.4                                                            & \multicolumn{1}{c|}{0.3}                                                            & 0.2                                                            & 0.1                                                            \\ \hline
        L2                      & \begin{tabular}[c]{@{}c@{}}51.1\\ \textcolor{gray}{(77.0)}\end{tabular}          & \begin{tabular}[c]{@{}c@{}}74.8\\ \textcolor{gray}{(89.5)}\end{tabular}          & \begin{tabular}[c]{@{}c@{}}\underline{46.1}\\ \textcolor{gray}{(68.9)}\end{tabular}          & \begin{tabular}[c]{@{}c@{}}\underline{67.3}\\ \textcolor{gray}{(80.7)}\end{tabular}          & \begin{tabular}[c]{@{}c@{}}49.8\\ \textcolor{gray}{(81.4)}\end{tabular}          & \begin{tabular}[c]{@{}c@{}}\underline{73.9}\\ \textcolor{gray}{(92.0)}\end{tabular}          & \begin{tabular}[c]{@{}c@{}}\underline{45.9}\\ \textcolor{gray}{(73.9)}\end{tabular}          & \begin{tabular}[c]{@{}c@{}}\underline{67.6}\\ \textcolor{gray}{(86.1)}\end{tabular}          & \begin{tabular}[c]{@{}c@{}}29.8\\ \textcolor{gray}{(64.2)}\end{tabular}          & \begin{tabular}[c]{@{}c@{}}47.9\\ \textcolor{gray}{(75.7)}\end{tabular}          & \begin{tabular}[c]{@{}c@{}}40.1\\ \textcolor{gray}{(79.6)}\end{tabular}          & \begin{tabular}[c]{@{}c@{}}69.1\\ \textcolor{gray}{(90.6)}\end{tabular}          & \begin{tabular}[c]{@{}c@{}}23.0\\ \textcolor{gray}{(64.6)}\end{tabular}          & \begin{tabular}[c]{@{}c@{}}39.5\\ \textcolor{gray}{(76.9)}\end{tabular}          & \begin{tabular}[c]{@{}c@{}}31.7\\ \textcolor{gray}{(83.0)}\end{tabular}          & \begin{tabular}[c]{@{}c@{}}58.0\\ \textcolor{gray}{(92.4)}\end{tabular}          & \begin{tabular}[c]{@{}c@{}}24.3\\ \textcolor{gray}{(47.1)}\end{tabular}          & \multicolumn{1}{c|}{\begin{tabular}[c]{@{}c@{}}40.2\\ \textcolor{gray}{(63.0)}\end{tabular}}          & \begin{tabular}[c]{@{}c@{}}22.2\\ \textcolor{gray}{(50.1)}\end{tabular}          & \begin{tabular}[c]{@{}c@{}}41.0\\ \textcolor{gray}{(74.0)}\end{tabular}          \\
        Influence \cite{BCT_shen2020towards}              & \begin{tabular}[c]{@{}c@{}}46.7\\ \textcolor{gray}{(81.9)}\end{tabular}          & \begin{tabular}[c]{@{}c@{}}71.8\\ \textcolor{gray}{(\underline{92.8})}\end{tabular}          & \begin{tabular}[c]{@{}c@{}}41.7\\ \textcolor{gray}{(70.9)}\end{tabular}          & \begin{tabular}[c]{@{}c@{}}64.8\\ \textcolor{gray}{(83.7)}\end{tabular}          & \begin{tabular}[c]{@{}c@{}}32.5\\ \textcolor{gray}{(82.3)}\end{tabular}          & \begin{tabular}[c]{@{}c@{}}53.8\\ \textcolor{gray}{(92.6)}\end{tabular}          & \begin{tabular}[c]{@{}c@{}}25.9\\ \textcolor{gray}{(64.8)}\end{tabular}          & \begin{tabular}[c]{@{}c@{}}48.4\\ \textcolor{gray}{(85.4)}\end{tabular}          & \begin{tabular}[c]{@{}c@{}}32.9\\ \textcolor{gray}{(70.8)}\end{tabular}          & \begin{tabular}[c]{@{}c@{}}56.8\\ \textcolor{gray}{(82.0)}\end{tabular}          & \begin{tabular}[c]{@{}c@{}}39.5\\ \textcolor{gray}{(79.9)}\end{tabular}          & \begin{tabular}[c]{@{}c@{}}67.5\\ \textcolor{gray}{(91.9)}\end{tabular}          & \begin{tabular}[c]{@{}c@{}}28.5\\ \textcolor{gray}{(75.6)}\end{tabular}          & \begin{tabular}[c]{@{}c@{}}52.0\\ \textcolor{gray}{(87.2)}\end{tabular}          & \begin{tabular}[c]{@{}c@{}}35.3\\ \textcolor{gray}{(85.3)}\end{tabular}          & \begin{tabular}[c]{@{}c@{}}61.7\\ \textcolor{gray}{(94.4)}\end{tabular}          & \begin{tabular}[c]{@{}c@{}}\underline{34.5}\\ \textcolor{gray}{(\underline{55.8})}\end{tabular}          & \multicolumn{1}{c|}{\begin{tabular}[c]{@{}c@{}}\underline{53.9}\\ \textcolor{gray}{(\underline{71.0})}\end{tabular}}          & \begin{tabular}[c]{@{}c@{}}\underline{34.5}\\ \textcolor{gray}{(\underline{59.9})}\end{tabular}          & \begin{tabular}[c]{@{}c@{}}\underline{56.7}\\ \textcolor{gray}{(\underline{79.4})}\end{tabular}          \\
        MMD \cite{DAN_Long}                    & \begin{tabular}[c]{@{}c@{}}42.5\\ \textcolor{gray}{(59.8)}\end{tabular}          & \begin{tabular}[c]{@{}c@{}}74.6\\ \textcolor{gray}{(78.5)}\end{tabular}          & \begin{tabular}[c]{@{}c@{}}42.9\\ \textcolor{gray}{(36.2)}\end{tabular}          & \begin{tabular}[c]{@{}c@{}}66.0\\ \textcolor{gray}{(52.3)}\end{tabular}          & \begin{tabular}[c]{@{}c@{}}49.0\\ \textcolor{gray}{(57.5)}\end{tabular}          & \begin{tabular}[c]{@{}c@{}}73.1\\ \textcolor{gray}{(74.6)}\end{tabular}          & \begin{tabular}[c]{@{}c@{}}42.9\\ \textcolor{gray}{(37.7)}\end{tabular}          & \begin{tabular}[c]{@{}c@{}}66.1\\ \textcolor{gray}{(55.0)}\end{tabular}          & \begin{tabular}[c]{@{}c@{}}37.7\\ \textcolor{gray}{(46.6)}\end{tabular}          & \begin{tabular}[c]{@{}c@{}}\underline{64.0}\\ \textcolor{gray}{(60.7)}\end{tabular}          & \begin{tabular}[c]{@{}c@{}}42.5\\ \textcolor{gray}{(61.7)}\end{tabular}          & \begin{tabular}[c]{@{}c@{}}\underline{74.6}\\ \textcolor{gray}{(79.9)}\end{tabular}          & \begin{tabular}[c]{@{}c@{}}36.6\\ \textcolor{gray}{(48.3)}\end{tabular}          & \begin{tabular}[c]{@{}c@{}}\underline{64.1}\\ \textcolor{gray}{(64.8)}\end{tabular}          & \begin{tabular}[c]{@{}c@{}}37.4\\ \textcolor{gray}{(59.1)}\end{tabular}          & \begin{tabular}[c]{@{}c@{}}\underline{67.4}\\ \textcolor{gray}{(77.7)}\end{tabular}          & \begin{tabular}[c]{@{}c@{}}16.2\\ \textcolor{gray}{(38.0)}\end{tabular}          & \multicolumn{1}{c|}{\begin{tabular}[c]{@{}c@{}}27.4\\ \textcolor{gray}{(52.9)}\end{tabular}}          & \begin{tabular}[c]{@{}c@{}}15.5\\ \textcolor{gray}{(35.1)}\end{tabular}          & \begin{tabular}[c]{@{}c@{}}27.0\\ \textcolor{gray}{(59.6)}\end{tabular}          \\ 
        Triplet \cite{hermans2017defense}                 & \begin{tabular}[c]{@{}c@{}} \underline{53.1} \\ \textcolor{gray}{(\underline{82.6})}\end{tabular}          & \begin{tabular}[c]{@{}c@{}}\underline{75.4}\\ \textcolor{gray}{(\underline{92.8})}\end{tabular}          & \begin{tabular}[c]{@{}c@{}}44.2\\ \textcolor{gray}{(\underline{71.9})}\end{tabular}          & \begin{tabular}[c]{@{}c@{}}64.8\\ \textcolor{gray}{(\underline{84.4})}\end{tabular}          & \begin{tabular}[c]{@{}c@{}}\underline{51.4}\\ \textcolor{gray}{(\underline{86.1})}\end{tabular}          & \begin{tabular}[c]{@{}c@{}}73.8\\ \textcolor{gray}{(\underline{93.9})}\end{tabular}          & \begin{tabular}[c]{@{}c@{}}  44.8 \\ \textcolor{gray}{(\underline{76.4})}\end{tabular}          & \begin{tabular}[c]{@{}c@{}}66.2\\ \textcolor{gray}{(\underline{87.5})}\end{tabular}          & \begin{tabular}[c]{@{}c@{}}
        \underline{40.5}\\ \textcolor{gray}{(\underline{71.8})}\end{tabular}          & \begin{tabular}[c]{@{}c@{}}62.1\\ \textcolor{gray}{(\underline{83.9})}\end{tabular}          & \begin{tabular}[c]{@{}c@{}}\underline{48.6}\\ \textcolor{gray}{(\underline{84.5})}\end{tabular}          & \begin{tabular}[c]{@{}c@{}}70.5\\ \textcolor{gray}{(\underline{93.5})}\end{tabular}          & \begin{tabular}[c]{@{}c@{}}\underline{39.8}\\ \textcolor{gray}{(\underline{77.6})}\end{tabular}          & \begin{tabular}[c]{@{}c@{}}61.1\\ \textcolor{gray}{(\underline{88.6})}\end{tabular}          & \begin{tabular}[c]{@{}c@{}}\underline{46.1}\\ \textcolor{gray}{(\underline{87.8})}\end{tabular}          & \begin{tabular}[c]{@{}c@{}}67.1\\ \textcolor{gray}{(\underline{94.8})}\end{tabular}          & \begin{tabular}[c]{@{}c@{}}29.0\\ \textcolor{gray}{(48.8)}\end{tabular}          & \multicolumn{1}{c|}{\begin{tabular}[c]{@{}c@{}}47.3\\ \textcolor{gray}{(65.2)}\end{tabular}}          & \begin{tabular}[c]{@{}c@{}}27.5\\ \textcolor{gray}{(50.3)}\end{tabular}          & \begin{tabular}[c]{@{}c@{}}50.1\\ \textcolor{gray}{(74.6)}\end{tabular}          \\
        RBCL (ours)              & \textbf{\begin{tabular}[c]{@{}c@{}}61.3\\ \textcolor{gray}{(85.0)}\end{tabular}} & \textbf{\begin{tabular}[c]{@{}c@{}}82.7\\ \textcolor{gray}{(93.7)}\end{tabular}} & \textbf{\begin{tabular}[c]{@{}c@{}}53.2\\ \textcolor{gray}{(74.3)}\end{tabular}} & \textbf{\begin{tabular}[c]{@{}c@{}}75.0\\ \textcolor{gray}{(85.5)}\end{tabular}} & \textbf{\begin{tabular}[c]{@{}c@{}}60.8\\ \textcolor{gray}{(88.0)}\end{tabular}} & \textbf{\begin{tabular}[c]{@{}c@{}}82.8\\ \textcolor{gray}{(95.0)}\end{tabular}} & \textbf{\begin{tabular}[c]{@{}c@{}}54.0\\ \textcolor{gray}{(79.5)}\end{tabular}} & \textbf{\begin{tabular}[c]{@{}c@{}}75.3\\ \textcolor{gray}{(89.8)}\end{tabular}} & \textbf{\begin{tabular}[c]{@{}c@{}}51.2\\ \textcolor{gray}{(75.6)}\end{tabular}} & \textbf{\begin{tabular}[c]{@{}c@{}}73.5\\ \textcolor{gray}{(86.4)}\end{tabular}} & \textbf{\begin{tabular}[c]{@{}c@{}}58.4\\ \textcolor{gray}{(85.4)}\end{tabular}} & \textbf{\begin{tabular}[c]{@{}c@{}}81.6\\ \textcolor{gray}{(93.8)}\end{tabular}} & \textbf{\begin{tabular}[c]{@{}c@{}}50.6\\ \textcolor{gray}{(79.7)}\end{tabular}} & \textbf{\begin{tabular}[c]{@{}c@{}}73.7\\ \textcolor{gray}{(90.2)}\end{tabular}} & \textbf{\begin{tabular}[c]{@{}c@{}}58.0\\ \textcolor{gray}{(88.5)}\end{tabular}} & \textbf{\begin{tabular}[c]{@{}c@{}}80.2\\ \textcolor{gray}{(95.4)}\end{tabular}} & \textbf{\begin{tabular}[c]{@{}c@{}}40.5\\ \textcolor{gray}{(60.6)}\end{tabular}} & \multicolumn{1}{c|}{\textbf{\begin{tabular}[c]{@{}c@{}}63.0\\ \textcolor{gray}{(75.5)}\end{tabular}}} & \textbf{\begin{tabular}[c]{@{}c@{}}39.5\\ \textcolor{gray}{(64.9)}\end{tabular}} & \textbf{\begin{tabular}[c]{@{}c@{}}64.8\\ \textcolor{gray}{(83.6)}\end{tabular}} \\ \hline
        UB                      & 85.8                                                           & 93.8                                                           & 76.6                                                           & 86.3                                                           & 89.1                                                           & 95.7                                                           & 81.0                                                           & 90.6                                                           & 76.6                                                           & 86.3                                                           & 85.8                                                           & 93.8                                                           & 81.0                                                           & 90.6                                                           & 89.1                                                           & 95.7                                                           & 58.7                                                           & \multicolumn{1}{c|}{74.5}                                                           & 67.6                                                           & 85.0                                                           \\ \hline
        \end{tabular}
        
        \caption{The experimental results for the comparison between our proposed RBCL and other methods. 
         The gray number in between ``()'' is the performance of the self-test retrieval. Our RBCL outperforms others in both cross-model and self-test retrieval by large margins. The best and second best performance for cross-model and self-test retrieval are in bold and underlined, respectively. }
        \label{table:all}
        \end{table*}
\begin{table*}[t]
\centering
\footnotesize
 \setlength\tabcolsep{3pt}
  \begin{tabular}{l|cc|cc|cc|cc|cc|cc|cc|cc|cccc}
\hline
                        & \multicolumn{4}{c|}{ID-S-1}                         & \multicolumn{4}{c|}{ID-S-2}                         & \multicolumn{4}{c|}{CD-S-1}                         & \multicolumn{4}{c|}{CD-S-2}                         & \multicolumn{4}{c}{CD-US}                                   \\ \hline
\multirow{2}{*}{Method} & \multicolumn{2}{c|}{M2M} & \multicolumn{2}{c|}{D2D} & \multicolumn{2}{c|}{M2M} & \multicolumn{2}{c|}{D2D} & \multicolumn{2}{c|}{M2D} & \multicolumn{2}{c|}{D2M} & \multicolumn{2}{c|}{M2D} & \multicolumn{2}{c|}{D2M} & \multicolumn{2}{c|}{M2D}          & \multicolumn{2}{c}{D2M} \\ \cline{2-21} 
                        & mAP         & R-1        & mAP         & R-1        & mAP         & R-1        & mAP         & R-1        & mAP         & R-1        & mAP         & R-1        & mAP         & R-1        & mAP         & R-1        & mAP  & \multicolumn{1}{c|}{R-1}  & mAP         & R-1        \\ \hline
baseline                    & 57.0        & 79.5       & 50.2        & 72.0       & 53.4        & 77.1       & 46.7        & 69.1       & 44.6        & 67.0       & 52.4        & 76.2       & 42.8        & 65.8       & 50.1        & 73.5       & 36.5 & \multicolumn{1}{c|}{57.6} & 35.0        & 59.6       \\
+ DGR              & 57.5        & 79.9       & 50.6        & 72.9       & 55.2        & 78.5       & 48.8        & 71.3       & 45.2        & 67.7       & 52.8        & 76.6       & 44.0        & 66.8       & 51.6        & 75.3       & 37.8    & \multicolumn{1}{c|}{58.9}    & 36.8           & 62.7          \\
+ NCAs              & 60.4        & 81.9       & 52.1        & 73.7       & 59.4        & 81.7       & 52.2        & 73.6       & 49.7        & 72.0       & 57.6        & 80.3       & 49.4        & 71.6       & 56.6        & 79.1       & 39.6 & \multicolumn{1}{c|}{62.0} & 38.9        & 63.4       \\

 + NCAs + DGR (ours)         & 61.3        & 82.7       & 53.2        & 75.0       & 60.8        & 82.8       & 54.0        & 75.3       & 51.2        & 73.5       & 58.4        & 81.6       & 50.6        & 73.7       & 58.0        & 80.2       & 40.5 & \multicolumn{1}{c|}{63.0} & 39.5        & 64.8       \\ \hline
\end{tabular}
    
\caption{The ablation of NCAs and DGR in RBCL. Cross-model retrieval performance is reported.}
\label{table:ablation of NAP}
\end{table*}

\section{Experimental Results}
In this section, we first complement some experimental details. Then,  we conduct primary experiments for all the settings. After that, we compare RBCL with baseline methods in the above mentioned five setting types. Finally, we analyze the effectiveness of RBCL with the ablation study and visualization.

\subsection{Experimental Details}
    For all the supervised learning settings, we conduct experiments based on BoT baseline         \cite{Luo_2019_CVPR_Workshops,Luo_2019_Strong_TMM}, and all the tricks and hyper-parameters are kept except that we remove the Center Loss \cite{wen2016discriminativecenterloss}. 
    
    For the unsupervised setting (CD-US), we use the baseline modified based on the open sourced UDA strong baseline \footnote{https://github.com/zkcys001/UDAStrongBaseline}. 
    To be consistent with BoT, we remove the non-local block and generalized mean pooling from the default setting and use the basic ResNet-50 as the backbone. We conduct the backward compatible training together with the UDA iteration. We update the labels of old features with the pseudo labels after each time of clustering. 
    

    The new model is \textbf{initialized} with the parameters of old model \textbf{by default} for all the reported results, except when the network structure changes (i.e., ID-S-2 and CD-S-2). We calculate $\mathcal{L}_{m}$ on the features before classifier with cosine similarity.  We add the DGR after the convergence of training loss. The mAP and Rank-1 accuracy are reported as evaluation metrics. Considering the variance between supervised and unsupervised learning, we set $K=100, \alpha=0.5$ for supervised settings and $K=40, \alpha=0.3$ for the unsupervised setting.  

\subsection{Primary Experiment}

        We first conduct primary experiments under all the settings  without adding any loss for improving the backward compatibility. The results are illustrated in Table. \ref{table:basic_ablation}. By observing the table, we summarize the conclusions as follows: 
        
        (1) The extended cross-domain settings are more challenging than the in-domain settings. The performance of Direct under the cross-domain settings is significantly lower than that of the in-domain settings. For example, the mAP of Direct in ID-S-1-M2M is 37.5\%, whilst only 17.9\% for CD-S-1-M2D. This is caused by the domain gap between source domain and target domain, and achieving backward compatibility in such a chaos old feature space is  rather challenging.
        
        (2) The unsupervised setting is more difficult than the supervised settings. This can be observed by comparing between the LB of CD-US and CD-S-1. For the supervised setting CD-S-1-M2D, it can achieve 32.2\% mAP in LB, whilst the unsupervised setting CD-US-M2D only achieves 0.4\% mAP. This decrement may be caused by the unstable pseudo labeling process in UDA iterations. The label space is different after each time of clustering,  therefore the current feature space is gradually shifted away from the old feature space.

        (3) The structure variation will increase the difficulty of achieving backward compatibility. By comparing the performance between ID-S-1 and ID-S-2, CD-S-1 and CD-S-2, we find that the performance of new model itself (UB) is improved, while the cross-model retrieval performance (LB) is decreased a lot, e.g., the mAP decreases from 50.0\% in ID-S-1-M2M to 0.2\% in ID-S-2-M2M. Intuitively, different structures have different capacities and parameters, thus the final feature spaces are different.

 
\begin{figure}[!t]
\centering

\subfloat[]{\includegraphics[width=1.7in]{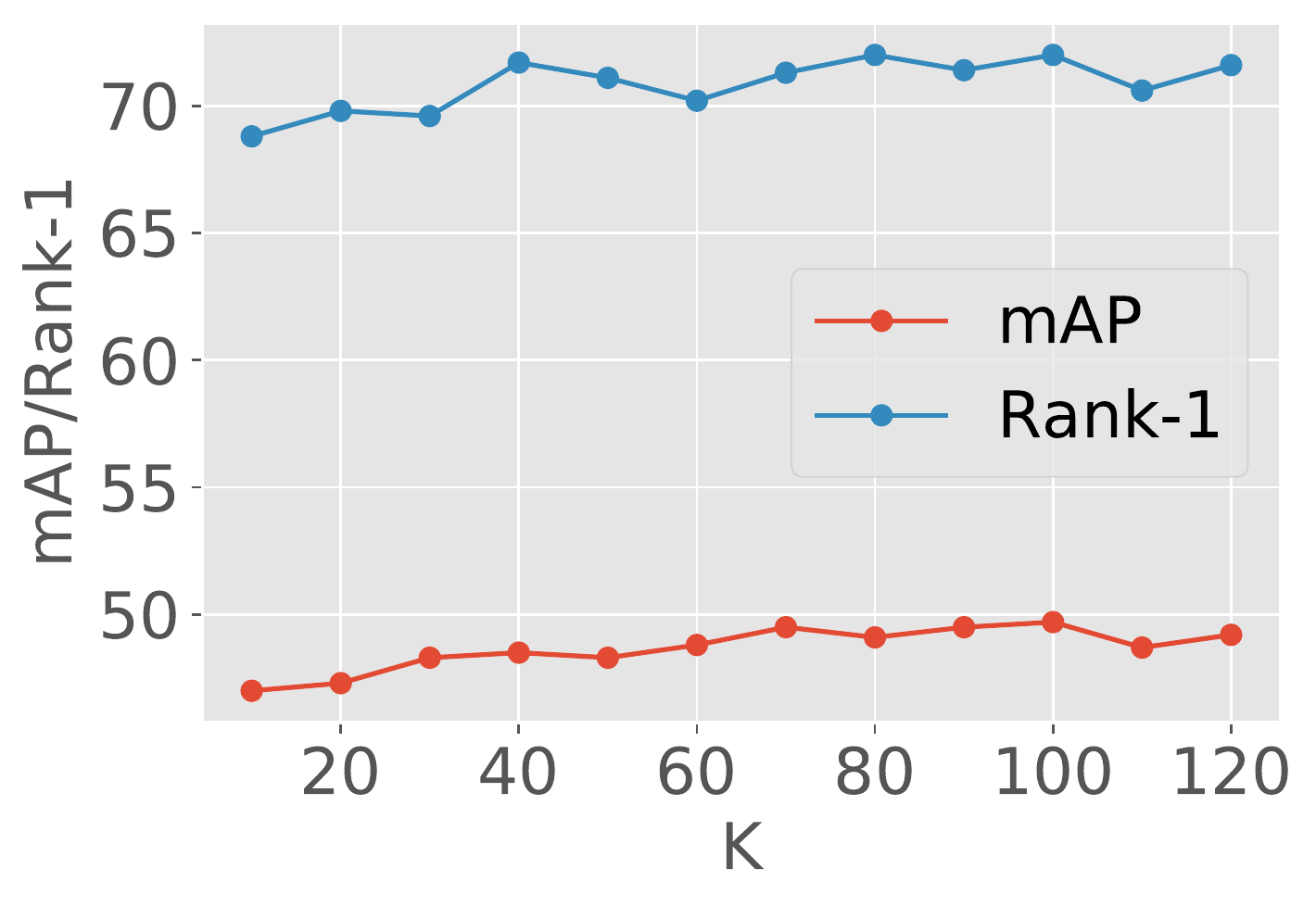}%
\label{fig_ablation_K}}
\subfloat[]{\includegraphics[width=1.7in]{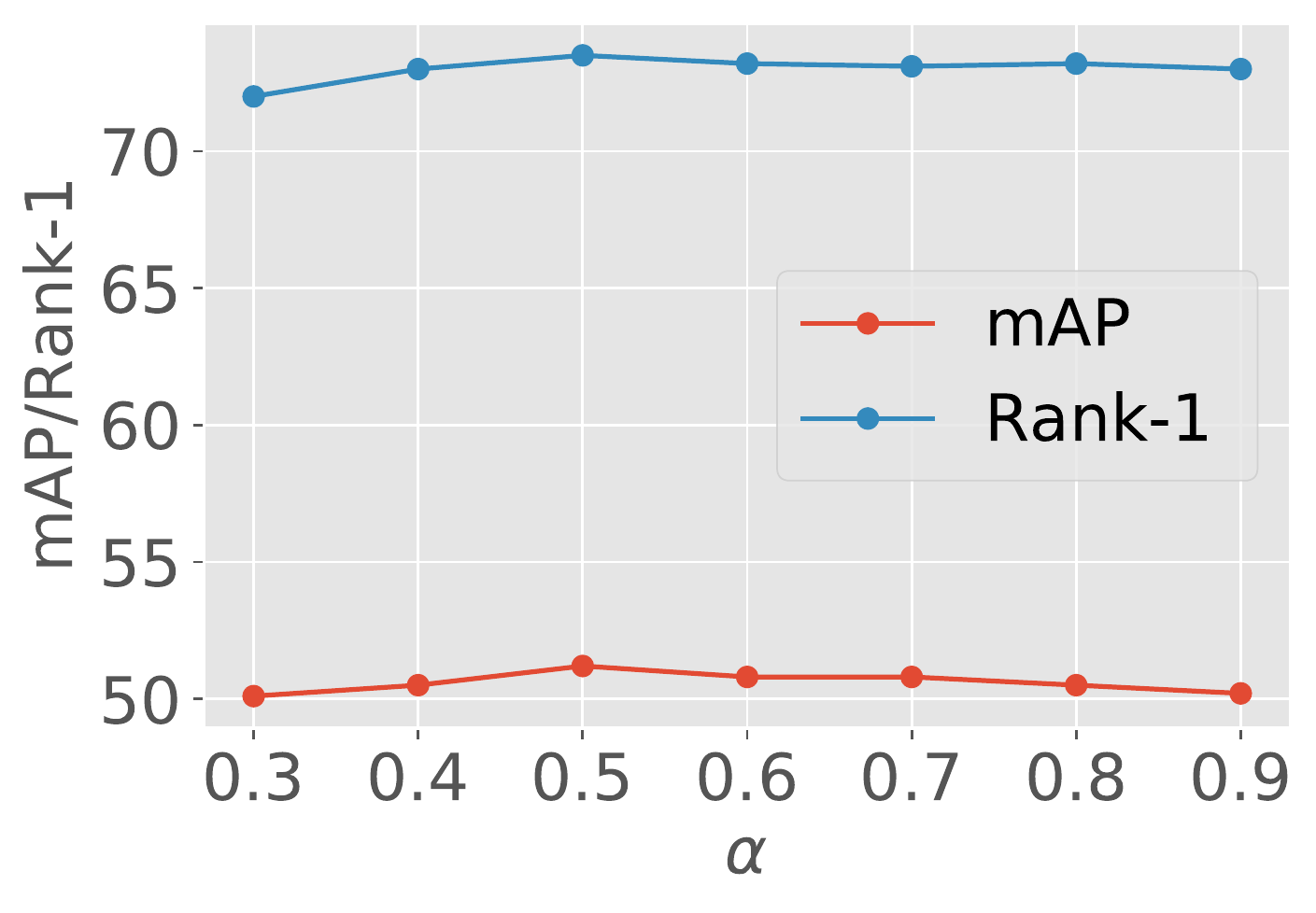}%
\label{fig_ablation_a}} 


\caption{(a) The influence of hyper-parameter $K$ in NCAs under CD-S-1-M2D. (b) The influence of hyper-parameter $\alpha$ in DGR under CD-S-1-M2D.}

\label{fig_ablation}
\end{figure}

\begin{figure*}[!t]
\centering
\subfloat[Distribution of triplets ($d_{nj}^i$).]{\includegraphics[width=2.2in]{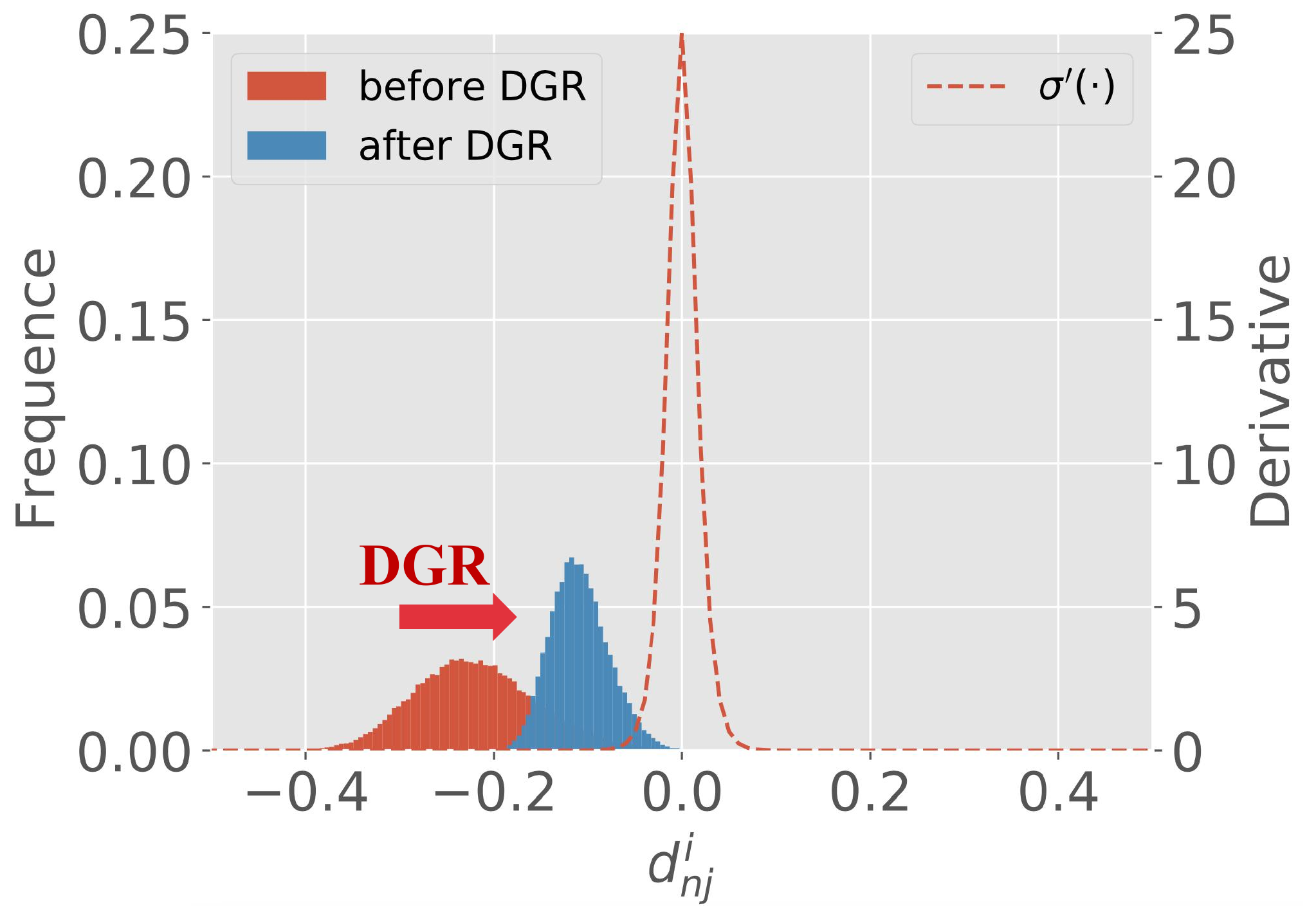}%
\label{fig_ditribution}} 
\subfloat[Gradient of triplets ($d_{nj}^i$).]{\includegraphics[width=2.0in]{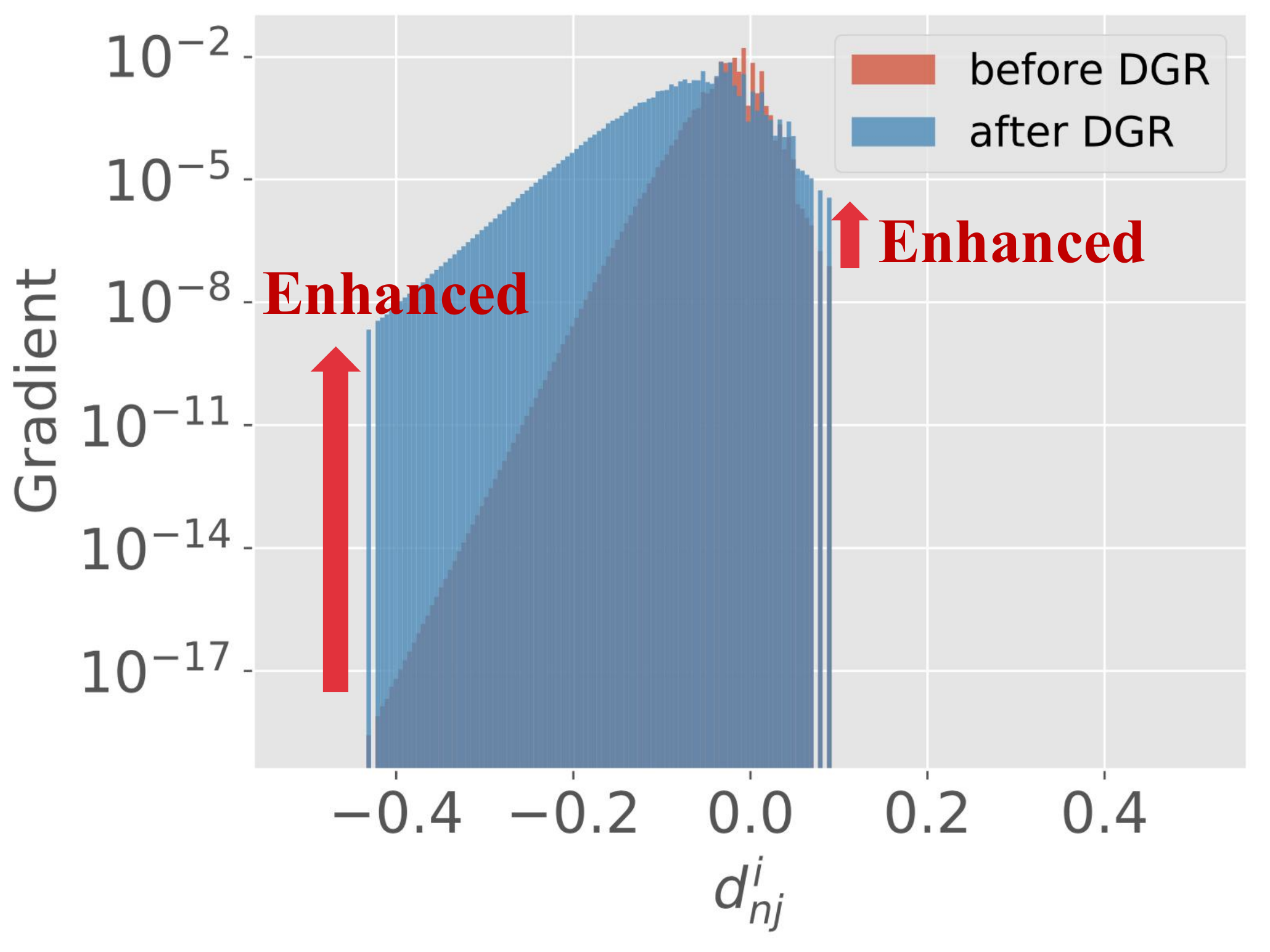}%
\label{fig_gradients}} 
\subfloat[Loss curve of $\mathcal{L}_m$ after adding DGR.]{\includegraphics[width=0.31\textwidth]{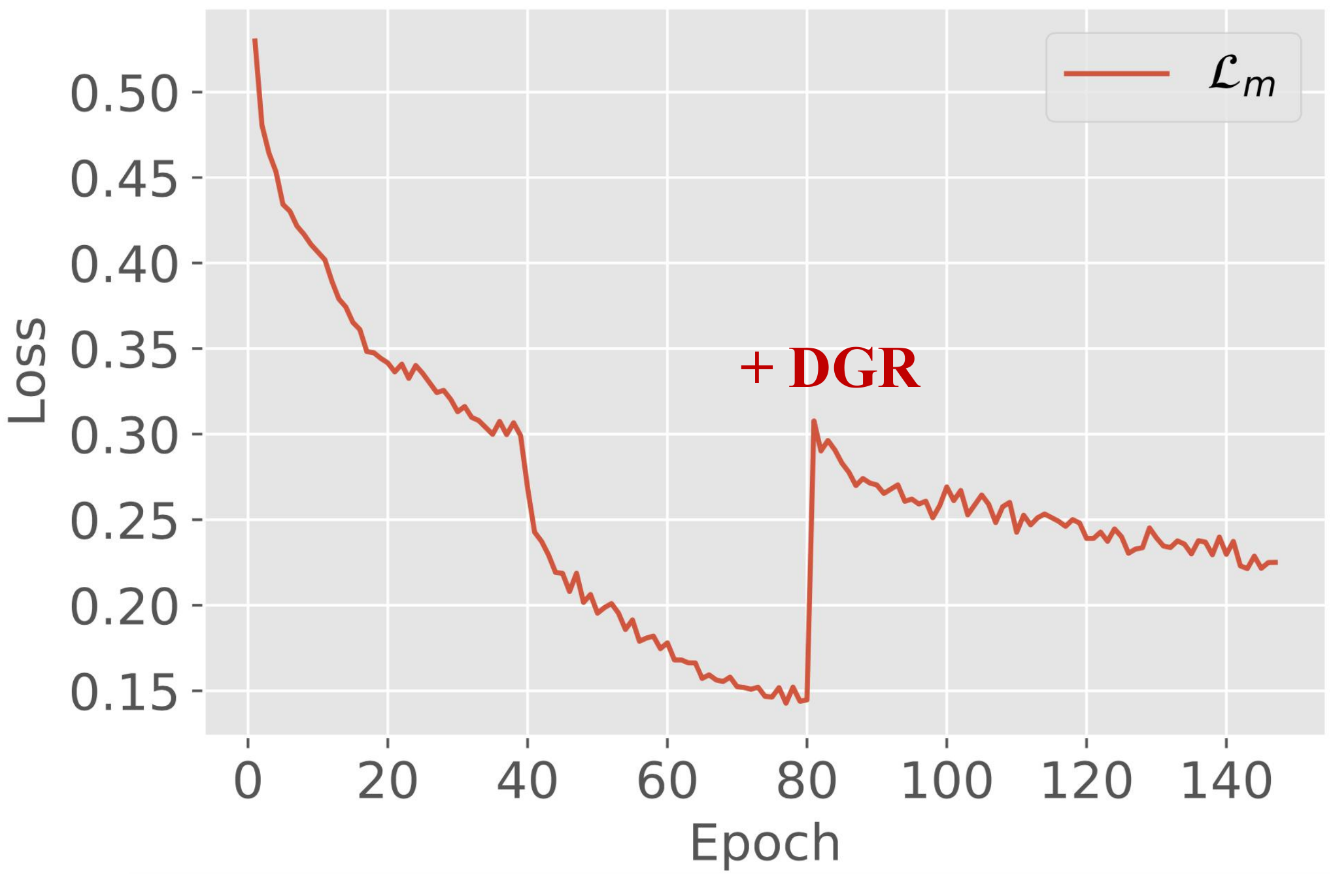}
\label{fig_loss_curve}}


\caption{ The visualization of distribution and gradient change of triplets ($d_{nj}^i$) in a mini-batch, and the loss curve of $\mathcal{L}_{m}$ (with NCAs) after adding DGR.  We add DGR after the convergence of $\mathcal{L}_m$ under the CD-S-1-M2D setting and $\alpha$ is set as $0.5$. 
(a) The distribution of triplets before (red) and after (blue) adding DGR, and the derivative of sigmoid function when $\tau=0.01$. 
(b) The average gradients in each interval before (red) and after (blue) adding DGR. We split the range of $d_{nj}^i$ into several intervals, and then report the average gradients for each interval. Note that the y axis is logarithmic. 
(c) The loss curve of $\mathcal{L}_{m}$ under CD-S-1-M2D. DGR is added after epoch 80.  }

\label{fig_DGR_visualization}
\end{figure*}
    
    \begin{figure}[!t]
    \centering
        
    

	\subfloat[Influence Loss]{\includegraphics[width=0.47\textwidth]{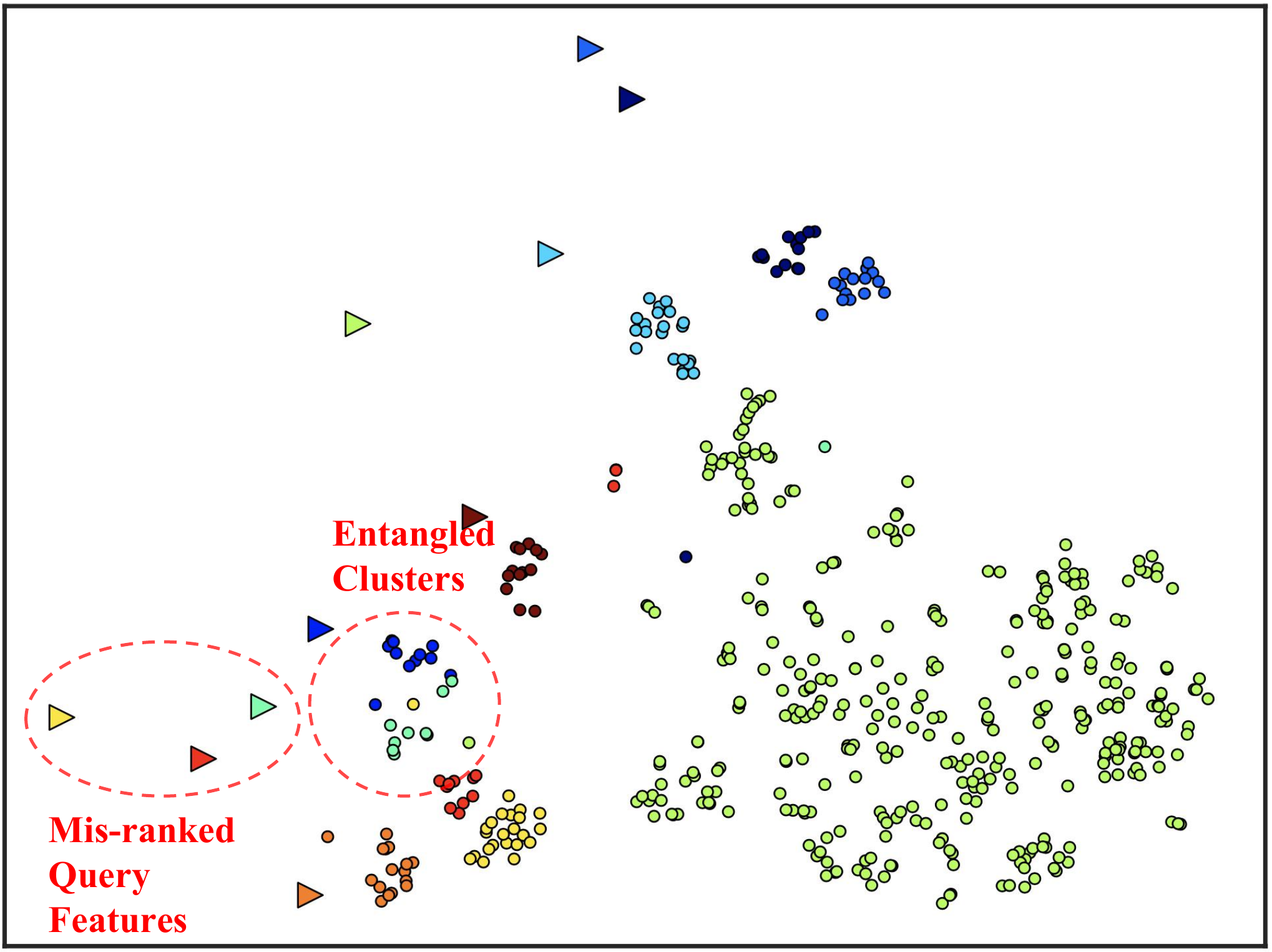} \label{fig_inf}}
    
	\subfloat[RBCL (ours)]{\includegraphics[width=0.47\textwidth]{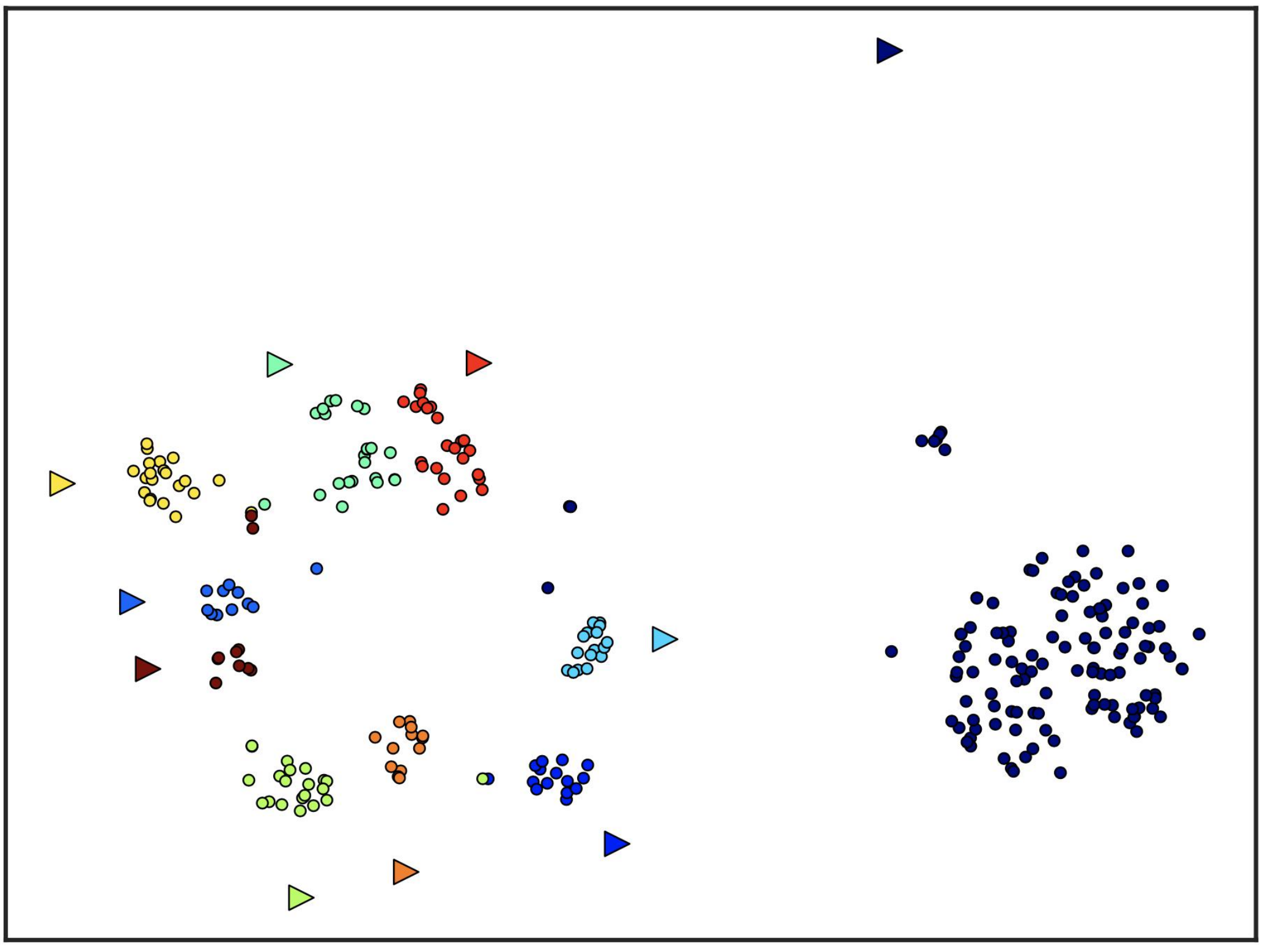}\label{fig_nap}} 

      \caption{The visualization of feature distributions for different methods under CD-US-M2D by t-SNE. The features of the first $10$ IDs of $\mathcal{D}_{D}$ are illustrated.  Triangles and circles represent for new query features and old gallery features, separately. Different colors denote for different IDs. Best view in color and zoom in. }
    	\label{fig:visualization_feature_space}
    	\vspace{0.2in}  
    \end{figure}

\subsection{Comparison with Baseline Methods}

    \subsubsection{Baseline Methods}
     The comparison between RBCL and baseline methods under all the settings is illustrated in Table. \ref{table:all}. The performance of  cross-model retrieval without additional loss for improving the backward compatibility  is reported as the baseline. 
     Aside from the cross-model retrieval performance, we also report the self-test performance of new model to illustrate the superiority of our RBCL.
     
     We compare our proposed RBCL with three distillation-based methods and one . The first one is to optimize the $L_2$ Loss between new features and the corresponding old ones  in a mini-batch. The second one is the Influence Loss proposed in \cite{BCT_shen2020towards}. The third one is the Maximum Mean Discrepancy (MMD) loss proposed in \cite{DAN_Long} which can make the distribution of new and old features in a mini-batch close to each other. The last one is Hard Mining Triplet Loss \cite{hermans2017defense}, which is also a typical ranking-based loss but belongs to the distance metric optimization.  

    \subsubsection{Analysis of Cross-model Performance}
    
    The cross-model retrieval performance of our RBCL outperforms other methods by large margins under all the settings.
    
    Compared with the distillation-based losses, in ID-S-1-D2D, RBCL outperforms Influence Loss by 11.7\% and 10.2\% in mAP and Rank-1, separately. 
    For Influence Loss and MMD Loss, their performance in ID-S-1 is even lower than the baseline, which significantly shows the inferiority of the distillation-based methods. They only push the new features to be close the old features, which will lead to mis-ranked new features, therefore the cross-model retrieval performance is inferior.
    
    Compared with another ranking-based method, in ID-S-2-M2M, RBCL surpasses Triplet Loss by 9.4\% and 9.0\% in mAP and Rank-1, respectively. In CD-S-1-M2D, RBCL surpasses it by 10.7\% and 11.3\% in mAP and Rank-1, respectively. The significant improvement shows that optimizing ranking metric is better than distance metric for BCT, and the proposed DGR and NCAs can help the new features achieve better ranking positions than the basic Triplet Loss.
    
    \subsubsection{Analysis of Self-test Performance}
    
    The self-test performance of RBCL is the closest one to the upper bound. By comparing the self-test performance of RBCL with the upper bound, we find that the backward compatible training process of RBCL has little impact on the performance of new model itself. For example, the Rank-1 in ID-S-1-M2M only decreases from 93.8\% to 93.7\%, and the mAP and Rank-1 in CD-US-M2D even outperform the upper bound by 1.9\% and 1.0\%, respectively. 
    This superiority is attributed to that RBCL only adjust the absolute position of  each cluster toward a best-ranking position, whilst the relative distribution between clusters are maintained, therefore the self-test performance will not decrease too much. 
    
    \subsubsection{Robustness to Structure Variation}
    
    By comparing the performance between ID-S-1 and ID-S-2, CD-S-1 and CD-S-2, we observe that the performance of RBCL is both high and robust in self-test and cross-model performance.  For the cross-model performance, the mAP difference and Rank-1 difference of RBCL between CD-S-1-M2D and CD-S-2-M2D are 0.6\% and 0.2\%, while for L2 Loss, the differences become 6.8\% and 8.4\%.
    As to the self-test performance, the difference between RBCL and upper bound is consistently small. For example, the Rank-1 difference in CD-S-1-D2M is 0.0\%. After the change of new model structure in CD-S-2-D2M, the difference merely increases by 0.3\%. When the structure changes, the gap between the new feature space and the old one becomes larger.  Distillation-based methods intend to roughly diminish the gap, whilst RBCL only tries to find better ranking order of samples, therefore the impact on self-test performance is smaller.

\subsection{Effectiveness of RBCL}

    \subsubsection{Influence of $K$ and $\alpha$}
    
     For supervised settings, we choose the hyper-parameters under CD-S-1-M2D, then adopt it for all the supervised settings.  
    
     For the choice of $K$ (Fig. \ref{fig_ablation_K}), we change $K$ sequentially from $10$ to $120$ with $10$ as the step size.  In general, the performance increases as $K$ becomes larger (with slight oscillation), which proves the effectiveness of NCAs. We set $K=100$ for all the supervised settings due to its best performance.
     
     For the ablation of $\alpha$ (Fig. \ref{fig_ablation_a}), we add DGR  after the convergence of using NCAs ($K=100$), and then adjust $\alpha$ from $0.3$ to $0.9$ with $0.1$ as the step size. Obviously, $\alpha=0.5$ gives the best performance because it provides the moderate gradients. 
     
    The similar ablation for unsupervised settings is conducted the same way, and we find that $K=40, \alpha=0.3$  works well.

    \subsubsection{Ablation of DGR and NCAs}
       
        The ablation of DGR and NCAs is illustrated in Table. \ref{table:ablation of NAP}. The ranking metric optimization between new features and the corresponding old features without DGR and NCAs is used as the baseline. We have the following important observations: 
        
        (1) DGR can boost the performance of both baseline method and the baseline+NCAs method under all the challenging settings. For example, 
        for the unsupervised setting CD-US-D2M, the Rank-1 increments are 3.1\% and 2.4\%, respectively. DGR solves the gradient vanish issue, which can refine the ranking positions and optimize the extremely hard negative samples, thus the performance is improved.
        
        (2) With the help of NCAs, the performance is largely boosted. After adding the NCAs to the baseline, for the in-domain setting ID-S-2-M2M,  the mAP is increased by 6.0\%. For the cross-domain settings CD-S-2-M2D and CD-S-2-D2M, the mAP are increased by 6.6\% and 6.5\%, separately. This strongly proves the effectiveness of our proposed NCAs, which successfully approximates the entire ole features space and brings the optimization with global perspective.
        
        (3) The combination of NCAs and DGR gives the best performance. 
        For example, the mAP and Rank-1 achieve increments of 7.8\%  and 7.9\% in CD-S-2-M2D, separately.
        For the challenging unsupervised setting CD-US-M2D, the mAP and Rank-1 are also increased by 4.0\% and 5.4\%, separately.
        
        In summary, both NCAs and DGR can boost the performance for all the challenging settings.

    \subsubsection{Visualization of DGR}
        
         We visualize the distribution and gradient changes in a mini-batch in Fig. \ref{fig_ditribution} and Fig. \ref{fig_gradients}. After adding DGR, the distribution is closer to the gradient-effective interval of sigmoid function (Fig. \ref{fig_ditribution}), therefore the gradients of the majority of triplets are exponentially enhanced (Fig. \ref{fig_gradients}). 
         We also illustrate the loss curve of $\mathcal{L}_{m}$ during the training of CD-S-1-M2D in Fig. \ref{fig_loss_curve}. We add DGR after epoch $80$. Obviously, the loss is enlarged after adding DGR, which means more ranking information is mined.


    \subsubsection{Visualization of Feature Space}
    
    To intuitively illustrate the principle difference between our RBCL and the existing distillation-based methods, we visualize the feature distributions in Fig. \ref{fig:visualization_feature_space}.
    We choose the Influence Loss as the representation of the distillation-based methods. We can easily observe that: (1) The old feature clusters are highly entangled with each other, which is due to the poor representation ability of the old model. (2) For the  Influence Loss, there exist quite a lot of  mis-ranked new query features, while for our RBCL, most of the new query features are in proper-ranking positions.

\section{Conclusion}
In this paper, we propose the effective RBCL for BCT task in person Re-ID. Different from the existing distillation-based methods, our proposed RBCL optimizes the ranking metric between new features and old features, which is superior in principle. To relieve the gradient vanish issue incurred by the sharp sigmoid function, we propose the DGR, which can reactivate the suppressed gradients. To further help targeting the global best-ranking positions, we propose to include NCAs during training, which can approximate the entire old feature space. RBCL outperforms other methods by a large margin under both in-domain and challenging cross-domain settings. In the future, we will study this problem on more retrieval tasks.


%



\section*{Acknowledgment}
This work was supported by the National Natural Science Foundation of China (No.62173302), the National Key Research and Development Program of China (2018YFE0119000) and the Fundamental Research Funds for the Central Universities (2021XZZX022).



 \bibliographystyle{elsarticle-num} 
 \bibliography{myref.bib}

\begin{thebibliography}{10}
\expandafter\ifx\csname url\endcsname\relax
  \def\url#1{\texttt{#1}}\fi
\expandafter\ifx\csname urlprefix\endcsname\relax\def\urlprefix{URL }\fi
\expandafter\ifx\csname href\endcsname\relax
  \def\href#1#2{#2} \def\path#1{#1}\fi

\bibitem{BCT_shen2020towards}
Y.~Shen, Y.~Xiong, W.~Xia, S.~Soatto, Towards backward-compatible
  representation learning, in: Proceedings of the IEEE/CVF Conference on
  Computer Vision and Pattern Recognition, 2020, pp. 6368--6377.

\bibitem{bct_wang2020unified}
C.-Y. Wang, Y.-L. Chang, S.-T. Yang, D.~Chen, S.-H. Lai, Unified representation
  learning for cross model compatibility, arXiv preprint arXiv:2008.04821
  (2020).

\bibitem{bct_chen2019r3}
K.~Chen, Y.~Wu, H.~Qin, D.~Liang, X.~Liu, J.~Yan, R3 adversarial network for
  cross model face recognition, in: Proceedings of the IEEE Conference on
  Computer Vision and Pattern Recognition, 2019, pp. 9868--9876.

\bibitem{UDA_fu2019self}
Y.~Fu, Y.~Wei, G.~Wang, Y.~Zhou, H.~Shi, T.~S. Huang, Self-similarity grouping:
  A simple unsupervised cross domain adaptation approach for person
  re-identification, in: Proceedings of the IEEE/CVF International Conference
  on Computer Vision, 2019, pp. 6112--6121.

\bibitem{UDA_ge2020MMT}
Y.~Ge, D.~Chen, H.~Li, Mutual mean-teaching: Pseudo label refinery for
  unsupervised domain adaptation on person re-identification, arXiv preprint
  arXiv:2001.01526 (2020).

\bibitem{UDA_ge2020SPL}
Y.~Ge, D.~Chen, F.~Zhu, R.~Zhao, H.~Li, Self-paced contrastive learning with
  hybrid memory for domain adaptive object re-id, arXiv preprint
  arXiv:2006.02713 (2020).

\bibitem{UDA_song2020unsupervised}
L.~Song, C.~Wang, L.~Zhang, B.~Du, Q.~Zhang, C.~Huang, X.~Wang, Unsupervised
  domain adaptive re-identification: Theory and practice, Pattern Recognition
  102 (2020) 107173.

\bibitem{UDA_ADCLUSTER_zhai2020ad}
Y.~Zhai, S.~Lu, Q.~Ye, X.~Shan, J.~Chen, R.~Ji, Y.~Tian, Ad-cluster: Augmented
  discriminative clustering for domain adaptive person re-identification, in:
  Proceedings of the IEEE/CVF Conference on Computer Vision and Pattern
  Recognition, 2020, pp. 9021--9030.

\bibitem{ye2016personranking}
M.~Ye, C.~Liang, Y.~Yu, Z.~Wang, Q.~Leng, C.~Xiao, J.~Chen, R.~Hu, Person
  reidentification via ranking aggregation of similarity pulling and
  dissimilarity pushing, IEEE Transactions on Multimedia 18~(12) (2016)
  2553--2566.

\bibitem{hermans2017defense}
A.~Hermans, L.~Beyer, B.~Leibe, In defense of the triplet loss for person
  re-identification, arXiv preprint arXiv:1703.07737 (2017).

\bibitem{DML_wang2019rankedlistloss}
X.~Wang, Y.~Hua, E.~Kodirov, G.~Hu, R.~Garnier, N.~M. Robertson, Ranked list
  loss for deep metric learning, in: Proceedings of the IEEE Conference on
  Computer Vision and Pattern Recognition, 2019, pp. 5207--5216.

\bibitem{DML_memorybankwang2020cross}
X.~Wang, H.~Zhang, W.~Huang, M.~R. Scott, Cross-batch memory for embedding
  learning, in: Proceedings of the IEEE/CVF Conference on Computer Vision and
  Pattern Recognition, 2020, pp. 6388--6397.

\bibitem{DML_Contrastive_oh2016deep}
H.~Oh~Song, Y.~Xiang, S.~Jegelka, S.~Savarese, Deep metric learning via lifted
  structured feature embedding, in: Proceedings of the IEEE conference on
  computer vision and pattern recognition, 2016, pp. 4004--4012.

\bibitem{DML_Contrastive_schroff2015facenet}
F.~Schroff, D.~Kalenichenko, J.~Philbin, Facenet: A unified embedding for face
  recognition and clustering, in: Proceedings of the IEEE conference on
  computer vision and pattern recognition, 2015, pp. 815--823.

\bibitem{DML_N_pair_sohn2016improved}
K.~Sohn, Improved deep metric learning with multi-class n-pair loss objective,
  in: Proceedings of the 30th International Conference on Neural Information
  Processing Systems, 2016, pp. 1857--1865.

\bibitem{DML_lift_structure_oh2016deep}
H.~Oh~Song, Y.~Xiang, S.~Jegelka, S.~Savarese, Deep metric learning via lifted
  structured feature embedding, in: Proceedings of the IEEE conference on
  computer vision and pattern recognition, 2016, pp. 4004--4012.

\bibitem{DML_proxy_NCA_movshovitz2017no}
Y.~Movshovitz-Attias, A.~Toshev, T.~K. Leung, S.~Ioffe, S.~Singh, No fuss
  distance metric learning using proxies, in: Proceedings of the IEEE
  International Conference on Computer Vision, 2017, pp. 360--368.

\bibitem{DML_brown2020smooth}
A.~Brown, W.~Xie, V.~Kalogeiton, A.~Zisserman, Smooth-ap: Smoothing the path
  towards large-scale image retrieval, arXiv preprint arXiv:2007.12163 (2020).

\bibitem{incremental_li2017Lwf}
Z.~Li, D.~Hoiem, Learning without forgetting, IEEE transactions on pattern
  analysis and machine intelligence 40~(12) (2017) 2935--2947.

\bibitem{meng2021lce}
Q.~Meng, C.~Zhang, X.~Xu, F.~Zhou, Learning compatible embeddings, in:
  Proceedings of the IEEE/CVF International Conference on Computer Vision,
  2021, pp. 9939--9948.

\bibitem{zheng2017discriminatively}
Z.~Zheng, L.~Zheng, Y.~Yang, A discriminatively learned cnn embedding for
  person reidentification, ACM Transactions on Multimedia Computing,
  Communications, and Applications (TOMM) 14~(1) (2017) 1--20.

\bibitem{marketzheng2015scalable}
L.~Zheng, L.~Shen, L.~Tian, S.~Wang, J.~Wang, Q.~Tian, Scalable person
  re-identification: A benchmark, in: Proceedings of the IEEE international
  conference on computer vision, 2015, pp. 1116--1124.

\bibitem{dukeristani2016performance}
E.~Ristani, F.~Solera, R.~Zou, R.~Cucchiara, C.~Tomasi, Performance measures
  and a data set for multi-target, multi-camera tracking, in: European
  Conference on Computer Vision, Springer, 2016, pp. 17--35.

\bibitem{dukezheng2017unlabeled}
Z.~Zheng, L.~Zheng, Y.~Yang, Unlabeled samples generated by gan improve the
  person re-identification baseline in vitro, in: Proceedings of the IEEE
  International Conference on Computer Vision, 2017, pp. 3754--3762.

\bibitem{he2016deepResnet}
K.~He, X.~Zhang, S.~Ren, J.~Sun, Deep residual learning for image recognition,
  in: Proceedings of the IEEE conference on computer vision and pattern
  recognition, 2016, pp. 770--778.

\bibitem{ibnpan2018two}
X.~Pan, P.~Luo, J.~Shi, X.~Tang, Two at once: Enhancing learning and
  generalization capacities via ibn-net, in: Proceedings of the European
  Conference on Computer Vision (ECCV), 2018, pp. 464--479.

\bibitem{DAN_Long}
M.~Long, Y.~Cao, J.~Wang, M.~I. Jordan,
  \href{http://jmlr.org/proceedings/papers/v37/long15.html}{Learning
  transferable features with deep adaptation networks}, in: Proceedings of the
  32nd International Conference on Machine Learning, {ICML} 2015, Lille,
  France, 6-11 July 2015, 2015, pp. 97--105.
\newline\urlprefix\url{http://jmlr.org/proceedings/papers/v37/long15.html}

\bibitem{Luo_2019_CVPR_Workshops}
H.~Luo, Y.~Gu, X.~Liao, S.~Lai, W.~Jiang, Bag of tricks and a strong baseline
  for deep person re-identification, in: The IEEE Conference on Computer Vision
  and Pattern Recognition (CVPR) Workshops, 2019.

\bibitem{Luo_2019_Strong_TMM}
H.~{Luo}, W.~{Jiang}, Y.~{Gu}, F.~{Liu}, X.~{Liao}, S.~{Lai}, J.~{Gu}, A strong
  baseline and batch normalization neck for deep person re-identification, IEEE
  Transactions on Multimedia (2019) 1--1\href
  {https://doi.org/10.1109/TMM.2019.2958756}
  {\path{doi:10.1109/TMM.2019.2958756}}.

\bibitem{wen2016discriminativecenterloss}
Y.~Wen, K.~Zhang, Z.~Li, Y.~Qiao, A discriminative feature learning approach
  for deep face recognition, in: European conference on computer vision,
  Springer, 2016, pp. 499--515.

\end{thebibliography}






\end{document}